\documentclass[lettersize,journal]{IEEEtran}

\usepackage{amsmath,amsfonts}
\usepackage{algorithmic}
\usepackage{array}
\usepackage[caption=false,font=normalsize,labelfont=rm,textfont=rm]{subfig}
\usepackage{textcomp}
\usepackage{stfloats}
\usepackage{url}
\usepackage{verbatim}
\usepackage{graphicx}
\def\BibTeX{{\rm B\kern-.05em{\sc i\kern-.025em b}\kern-.08em
    T\kern-.1667em\lower.7ex\hbox{E}\kern-.125emX}}
\usepackage{balance}

\usepackage{booktabs}

\usepackage{colortbl}    % for \rowcolor
\usepackage{xcolor}
\definecolor{myblue}{RGB}{238, 242, 254}

\definecolor{mygray}{gray}{.9}
\usepackage{multirow}
\usepackage{tablefootnote}
\definecolor{revised}{RGB}{0, 0, 0}
\definecolor{revised_checked}{RGB}{0, 0, 0}
\definecolor{revised_r1}{RGB}{0, 0, 0}

\usepackage{enumitem}
\usepackage[pagebackref=true,breaklinks=true,letterpaper=true,colorlinks,bookmarks=false]{hyperref}
\usepackage{amssymb}

\usepackage{amsthm}
\definecolor{markred}{RGB}{200,0,0}
\definecolor{commentblue}{RGB}{0,70,180}

\begin{document}

\title{StableMind: Source-Free Cross-Subject fMRI Decoding with Regularized Adaptation}

% \author{IEEE Publication Technology Department
% \thanks{Manuscript created October, 2020; This work was developed by the IEEE Publication Technology Department. This work is distributed under the \LaTeX \ Project Public License (LPPL) ( http://www.latex-project.org/ ) version 1.3. A copy of the LPPL, version 1.3, is included in the base \LaTeX \ documentation of all distributions of \LaTeX \ released 2003/12/01 or later. The opinions expressed here are entirely that of the author. No warranty is expressed or implied. User assumes all risk.}}

\author{Jintao Guo$^{\dagger}$, Lin Wang$^{\dagger}$, Shumeng Li, Jian Zhang, Yulin Zhou, Luyang Cao, Hairong Zheng$^*$, Yinghuan Shi$^*$ 

\thanks{
$\dagger$ Jintao Guo and Lin Wang contributed equally to this work.

$*$ Corresponding authors: Hairong Zheng and Yinghuan Shi.

Jintao Guo, Shumeng Li, Jian Zhang, Yulin Zhou, Luyang Cao, and Yinghuan Shi are with the National Key Laboratory for Novel Software Technology and the 
Institute of Brain-Machine Interface, Nanjing University, Nanjing 210023, China.
Jian Zhang is also with the School of Intelligence Science and Technology, Nanjing University, Nanjing 215163, China (e-mail: guojintao@smail.nju.edu.cn
; lism@smail.nju.edu.cn
; zhang.jian@nju.edu.cn
; zhouyulin@smail.nju.edu.cn
; caoluyang@smail.nju.edu.cn
; syh@nju.edu.cn).

Lin Wang is with the School of Electrical and Electronic Engineering, Nanyang Technological University, Singapore 639798 (e-mail: linwang@ntu.edu.sg).

Hairong Zheng is with the Institute of Brain-Machine Interface, Nanjing University, Nanjing 210023, China, and also with the Shenzhen Institute of Advanced Technology, Chinese Academy of Sciences, Shenzhen 518055, China (e-mail: hr.zheng@siat.ac.cn).
}}

\maketitle
\begin{abstract}
  Existing cross-subject fMRI decoding methods typically train a model on multiple scanned subjects and then adapt it to a new subject using substantial paired fMRI-image data. 
  However, in realistic scenarios, new-subject fMRI data are often limited due to costly data acquisition, and raw data from previous subjects may be inaccessible, leading existing methods to suffer performance degradation during new-subject adaptation.
  In this paper, we identify that this degradation stems from two key issues: brain-side instability caused by large subject differences in fMRI responses, and image-side supervision unreliability caused by fine-grained visual details that are not reliably supported by limited fMRI signals. 
  To address these challenges, we propose StableMind, a regularized adaptation framework designed to improve brain-side representation stability and image-side supervision reliability.
  (1) To stabilize brain representations, StableMind reuses ridge projections from the pretrained model as adaptation priors to constrain limited-data new-subject adaptation, and applies Fourier-based feature-level brain augmentation to improve robustness to individual variability.
  (2) To improve image supervision reliability, StableMind introduces difficulty-aware image blur for brain-image alignment, reducing the influence of fine-grained visual details that are weakly supported by limited fMRI signals while preserving stable visual structure.
  Experiments on the Natural Scenes Dataset under a unified 1-hour adaptation protocol demonstrate that StableMind achieves 84.02\% image retrieval accuracy and 81.66\% brain retrieval accuracy averaged over four subjects, surpassing the state-of-the-art method by 5.71\% brain retrieval accuracy with fewer trainable adaptation parameters.
  Our code is available at \textcolor{magenta}{\href{https://github.com/lingeringlight/StableMind}{https://github.com/lingeringlight/StableMind}}.
\end{abstract}


\section{Introduction}
\label{sec:intro}

\begin{figure}
  \centering
    \includegraphics[width=0.98\linewidth]{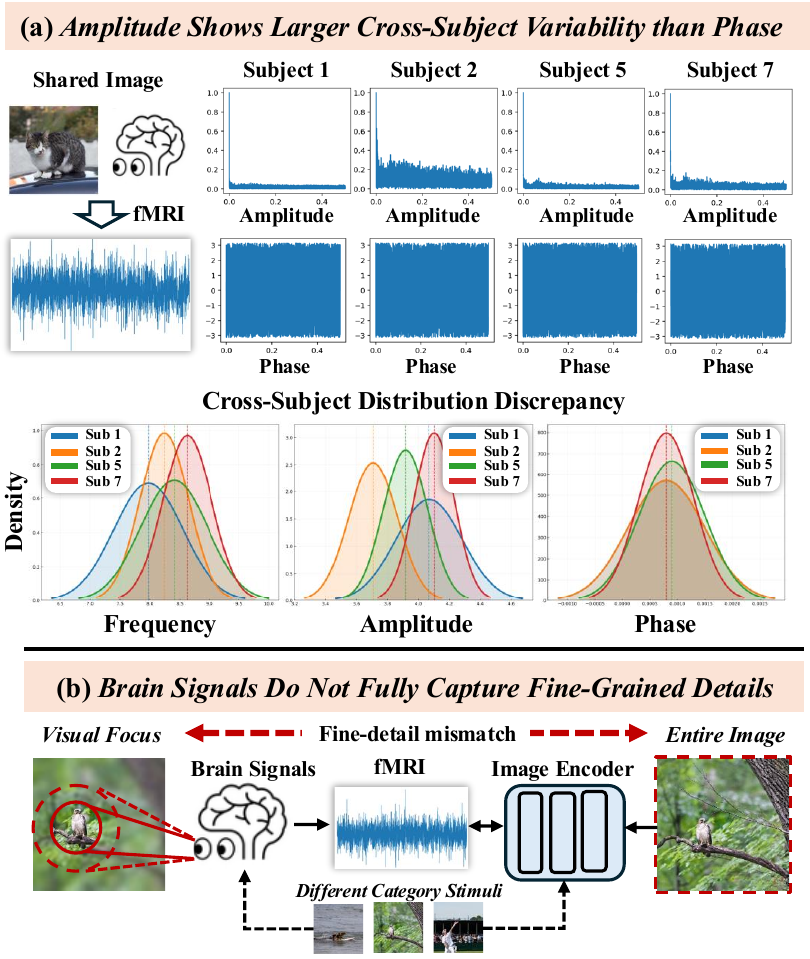}
   \caption{Two key challenges in cross-subject brain decoding. The experiments are conducted on images shared across $4$ subjects. 
   (a) Cross-subject fMRI representations exhibit noticeable subject-dependent variations, which make direct adaptation to a new subject prone to overfitting under limited data. 
   (b) A conceptual illustration of image-side mismatch under limited-data adaptation: full-image supervision may contain fine-grained details that are not always reliably reflected in brain responses. Directly enforcing such full-detail brain-image alignment may therefore introduce unreliable visual cues.
   }
  \label{fig: motivation}
\end{figure}

Understanding how the human brain encodes visual perception is a core problem in cognitive science and Brain-Computer Interface (BCI) research \cite{horikawa2017generic,qian2020binless}. Functional magnetic resonance imaging (fMRI) provides a noninvasive basis for this study by offering precise localization of functional regions in the visual cortex \cite{lin2019dcnn,beliy2019voxels}. Brain decoding aims to reconstruct natural visual stimuli from fMRI brain activity and to visualize internal mental representations \cite{mai2023unibrain,xia2024dream,chen2023seeing,scotti2023reconstructing}. Recent progress leverages large-scale generative and multimodal models by translating fMRI patterns into embeddings of pretrained deep networks and then performing image retrieval or reconstruction, often with vision-language priors such as CLIP \cite{radford2021learning} and diffusion models \cite{rombach2022high}.
This paradigm has substantially improved reconstruction quality and provided a useful framework for studying the correspondence between neural representations and visual semantics \cite{gong2025mindtuner,wang2024mindbridge}.

Early fMRI decoding studies typically adopt a subject-specific paradigm, where a separate decoder is trained for each individual subject to obtain high-fidelity reconstructions~\cite{scotti2023reconstructing,takagi2023high}. 
Although effective, this paradigm requires large amounts of fMRI recordings for every new subject and generalizes poorly across individuals due to substantial inter-subject variability in brain responses~\cite{scotti2024mindeye2,mai2023unibrain}. 
This severely limits the scalability of fMRI decoding in practical brain-computer interface and clinical scenarios, where collecting tens of hours of high-quality fMRI data for each new subject is often expensive and impractical. 
To alleviate this limitation, recent cross-subject decoding methods pretrain a shared model on multiple source subjects and adapt it to a new target subject with limited fMRI data~\cite{scotti2024mindeye2,wang2024mindbridge,dai2025mindaligner,gong2025mindtuner}. 
In realistic deployment, however, source-subject fMRI signals may be unavailable during target adaptation due to privacy, storage, or data-sharing constraints. 
This leads to a practical source-free cross-subject adaptation scenario, where a multi-subject pretrained decoder must be adapted to a new subject using only limited target-subject fMRI data.

In this work, we revisit source-free cross-subject fMRI decoding from a regularized adaptation perspective. 
We identify two coupled sources of instability in this setting. 
First, \textit{fMRI representations exhibit substantial subject-specific variations, even under shared stimuli and standardized acquisition protocols}.
As illustrated in Fig.~\ref{fig: motivation}(a), even for shared visual stimuli in NSD, brain representations from different subjects show noticeable distributional variations, especially in amplitude-related spectral statistics.
Under limited target-subject data, directly adapting the subject-specific voxel-to-latent projection may overfit to these subject-specific patterns, leading to unstable brain-to-visual mapping.
Second, \textit{the correspondence between fMRI representations and fine-grained visual details could be unstable under limited-data adaptation}. 
As shown in Fig.~\ref{fig: motivation}(b), full-image supervision preserves rich visual details, whereas the corresponding brain responses may provide only partial or noisy evidence for these details. 
Directly enforcing full-detail image supervision may thus introduce overly strong cross-modal constraints, guiding the model to fit visual cues that are not reliably supported by the limited target-subject fMRI data.
These factors suggest that the instability of source-free cross-subject fMRI decoding arises not only from subject-level distribution discrepancy, but also from stimulus-level mismatch between brain responses and image supervision.

To address these challenges, we propose \textbf{StableMind}, a regularized source-free adaptation framework for cross-subject fMRI decoding under limited target-subject data. 
Different from prior methods that mainly emphasize latent alignment or lightweight fine-tuning, StableMind explicitly stabilizes target adaptation at both the subject-transfer and cross-modal supervision levels. 
For \textbf{subject-transfer regularization}, StableMind introduces cross-subject ridge reuse, which incorporates source-subject projection priors into the target voxel-to-latent mapping and regularizes the under-constrained target projection during adaptation. 
To further improve representation robustness, StableMind employs Fourier-based feature-level brain augmentation, which perturbs amplitude-related statistics of intermediate brain features while preserving their structural phase information, reducing sensitivity to subject-dependent variations. 
For \textbf{cross-modal supervision regularization}, StableMind develops a difficulty-aware image blur strategy that adaptively modulates image-side supervision according to sample-level alignment reliability. 
This strategy reduces the influence of unreliable fine-grained details for easier samples while preserving sufficient visual structure for harder samples, leading to more reliable brain-image alignment under limited data.
In summary, our contributions are as follows:
\begin{itemize}[itemsep=2pt,topsep=2pt]
      \item We revisit source-free cross-subject fMRI decoding from a regularized adaptation perspective, identifying two coupled instability sources under limited target-subject data, \textit{i.e.}, subject-level fMRI distribution discrepancy and stimulus-level brain-image supervision mismatch.
      \item We propose StableMind, a regularized source-free adaptation framework that explicitly targets the two instability sources. It alleviates subject-level representation discrepancy via source-guided projection reuse and feature-level brain augmentation, and mitigates stimulus-level supervision mismatch via difficulty-aware image supervision.
      \item Extensive experiments on the NSD benchmark under the unified 1-hour adaptation protocol show that StableMind achieves superior retrieval performance and competitive reconstruction quality with fewer trainable adaptation parameters, \textit{e.g.}, improving brain retrieval by $5.71\%$ ($81.66\%$ vs. $75.95\%$) over the state-of-the-art method.
\end{itemize}

\section{Related Works}
\textbf{fMRI-Based Brain Decoding.} 
Brain decoding aims to reconstruct stimuli perceived by subjects from their brain activity, providing a promising tool for analyzing how the brain processes external information \cite{naselaris2011encoding,kim2024brain,mathis2024decoding}. 
Early studies \cite{horikawa2017generic,shen2019deep} revealed the layer-wise correspondences between visual cortex activity and deep neural networks (DNNs) representations, thus proposing to use linear mappings from fMRI to intermediate feature spaces to decode coarse visual attributes. 
With advances in deep generative models \cite{goodfellow2020generative,ho2020denoising}, brain decoding shifted from feature prediction to visual stimuli reconstruction, which is achieved by mapping brain signals to the latent spaces of large models \cite{scotti2023reconstructing,gao2024mind,gao2025mind}. 
Recently, brain decoding has been greatly advanced by multimodal methods, including self-supervised or masked brain modeling to denoise neural features \cite{chen2023seeing,bao2025wills}, contrastive learning to align fMRI and multimodal priors \cite{lin2022mind,xu2023versatile,quan2024psychometry}, and fine-grained supervision using multi-layer or token-level CLIP \cite{radford2021learning} features \cite{scotti2023reconstructing,chen2023seeing,wang2024mindbridge,xia2024dream}, as well as unified or language-augmented decoding systems such as UniBrain \cite{wang2024unibrain}, UMBRAE \cite{xia2024umbrae}, and NVL \cite{shen2024neuro}.
Despite advances, these models are trained independently per subject, requiring dozens of hours of expensive fMRI training data to attain high-quality results \cite{scotti2023reconstructing}.
Recent pioneering studies have explored cross-subject decoding by pretraining on multiple subjects and fine-tuning on a new subject with limited data \cite{scotti2024mindeye2,wang2024mindbridge,dai2025mindaligner,gong2025mindtuner,li2026duala}, but they mainly focus on representation alignment, leaving the stability of limited-data target adaptation less explicitly addressed.
In contrast, we propose a unified adaptation framework that jointly improves subject-specific input mapping, brain-side representation robustness, and image-side supervision during fine-tuning, enabling robust cross-subject decoding under limited data.

\textbf{Cross-Subject Functional Alignment.} 
In visual decoding, single-subject models have exposed the issue of excessive reliance on the data volume of individual subjects \cite{ozcelik2023natural,scotti2023reconstructing,takagi2023high}, leading researchers to shift toward cross-subject studies \cite{liu2025see,wang2024mindbridge,scotti2024mindeye2}. 
However, as brains differ both in size and processing mechanisms \cite{finn2017can,allen2022massive}, the resulting variability in fMRI signals has spurred research into brain alignment methods. 
Early studies typically depend on shared stimuli, requiring paired data from multiple subjects exposed to identical visual inputs, and learned functional mappings through reconstruction loss optimization \cite{rastegarnia2023brain}.
To remove this limitation, recent works have explored functional alignment in latent space \cite{scotti2023reconstructing,ferrante2024through,wang2024mindbridge,gong2025mindtuner}, which allows these methods to perform cross-subject visual decoding on the Natural Scenes Dataset (NSD) \cite{allen2022massive}, the largest open-source dataset that lacks shared stimuli. 
Representatively, MindEye2 \cite{scotti2024mindeye2} employs ridge regression to align different subjects into a shared latent space, followed by a shared decoding module to learn fMRI-to-stimuli mappings.
MindBridge \cite{wang2024mindbridge} proposes to generate pseudo shared stimuli to construct paired data for brain alignment.
% Furthermore, MindAligner \cite{dai2025mindaligner} introduces explicit functional alignment modules to reduce semantic conflicts across subjects, while 
MindTuner \cite{gong2025mindtuner} explores lightweight LoRA-style adapters for fine-tuning a shared backbone to improve data-efficient adaptation.
Duala \cite{li2026duala} further introduces dual-level alignment of subjects and stimuli to improve cross-subject fMRI decoding, but its adaptation process relies on leveraging source-subject distribution information.
Despite their success, these methods mainly focus on latent functional alignment or lightweight fine-tuning, with less attention to adaptation stability under limited new-subject data.
%  and inaccessible raw source-subject fMRI-image pairs. 
In contrast, StableMind stabilizes target-subject adaptation by jointly regularizing subject-specific voxel-to-latent mapping, brain-side representation learning, and image-side supervision, making it suitable for limited-data adaptation without revisiting raw source-subject fMRI-image pairs.

\section{Method}

\subsection{Problem Setting and Notations}
Collecting high-quality fMRI signals is costly and time-consuming, which makes brain decoding particularly challenging in limited-data adaptation settings \cite{scotti2024mindeye2}. In this work, we focus on source-free cross-subject brain decoding, where a model pretrained on multiple source subjects is adapted to a new target subject using only a small amount of target-subject fMRI data \cite{gong2025mindtuner,scotti2024mindeye2}. 
During this adaptation stage, raw source-subject fMRI-image pairs are unavailable.
Following prior settings \cite{scotti2024mindeye2}, we first pretrain a multi-subject model on full training sessions from source subjects and then fine-tune it on a held-out subject using only a single one-hour fMRI session, corresponding to approximately 2.5\% of the full subject-specific training data.
Formally, for each subject $s$, we denote the paired fMRI-image data as $(I, V^{(s)})$, where $I \in \mathbb{R}^{3 \times H \times W}$ is the visual stimulus and $V^{(s)} \in \mathbb{R}^{d_s}$ is the corresponding voxel response with $d_s$ voxels. A subject-specific ridge mapper $\mathbf{R}^{(s)}: \mathbb{R}^{d_s} \rightarrow \mathbb{R}^{K}$ projects voxel responses into a shared latent space, yielding
$
M = \mathbf{R}^{(s)}(V^{(s)}) \in \mathbb{R}^{K}.
$
The brain encoder $F(\cdot;\theta)$ then transforms $M$ into token-level embeddings $Z_B \in \mathbb{R}^{T \times C}$, which are aligned with CLIP image tokens $Z_V = F_V(I)$ extracted by a frozen image encoder $F_V$. A diffusion prior further predicts CLIP image embeddings from brain tokens, enabling end-to-end reconstruction. The baseline training objective follows MindEye2 and combines the diffusion-prior loss $\mathcal{L}_{\text{prior}}$, the low-level blurry reconstruction loss $\mathcal{L}_{\text{low}}$, and a bidirectional contrastive loss $\mathcal{L}_{\text{BiMixCo}}$.

\begin{figure*}
  \centering
    \includegraphics[width=\linewidth]{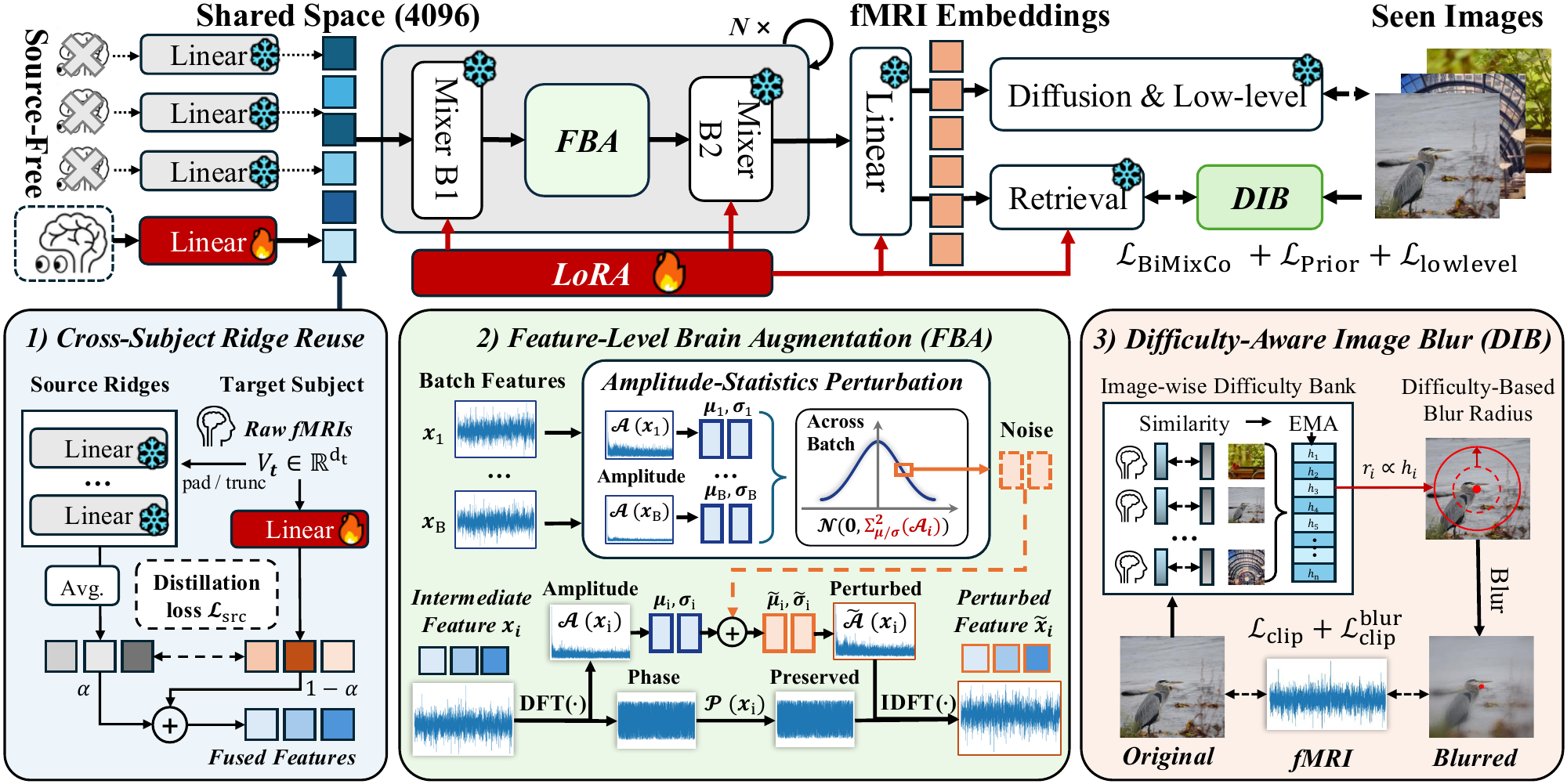}
  \caption{Overview of StableMind. StableMind targets cross-subject fMRI-to-image decoding under limited new-subject data. 
  (a) A multi-subject pretrained decoder is adapted to a new subject using only a single fMRI session. 
  (b) On the brain side, StableMind improves adaptation through \textit{cross-subject ridge reuse}, which injects source-subject priors into the subject-specific voxel-to-hidden mapping, and \textit{feature-level brain augmentation}, centered on Fourier-based spectral perturbation, which improves representation robustness under subject-specific variations.   
  (c) On the image side, StableMind introduces a \textit{difficulty-aware image blur} that adaptively modulates supervision according to sample-level alignment difficulty, reducing the influence of unreliable high-frequency details during cross-modal alignment. Together, these components improve the robustness of cross-subject brain-image alignment under limited-data adaptation.}
  \label{fig: framework}
\end{figure*}

Building upon this foundation, we focus on the new-subject fine-tuning stage, where only limited fMRI data from the target subject are available. 
We propose \textbf{StableMind}, a regularized source-free adaptation framework that stabilizes limited-data target adaptation from two aspects, \textit{i.e.}, subject-level representation regularization and stimulus-level supervision regularization.
% As shown in Fig.~\ref{fig: framework}, StableMind contains two regularization branches corresponding to the two instability sources. The subject-level branch stabilizes target brain representations through cross-subject ridge reuse and feature-level brain augmentation, while the stimulus-level branch improves supervision reliability through difficulty-aware image blur.
As shown in Fig.~\ref{fig: framework}, StableMind consists of three complementary modules that regularize different stages of adaptation: (1) \textit{cross-subject ridge reuse}, which incorporates source-subject priors into the voxel-to-hidden projection; (2) \textit{feature-level brain augmentation}, centered on Fourier-based spectral perturbation, which improves the robustness of learned brain representations; and (3) \textit{difficulty-aware image blur}, which adaptively reshapes image-side supervision by sample-level alignment difficulty. In the following parts, we describe each component in detail.

\subsection{Cross-Subject Ridge Reuse}

\begin{figure}[!tb]
   \centering
    \includegraphics[width=0.8\linewidth]{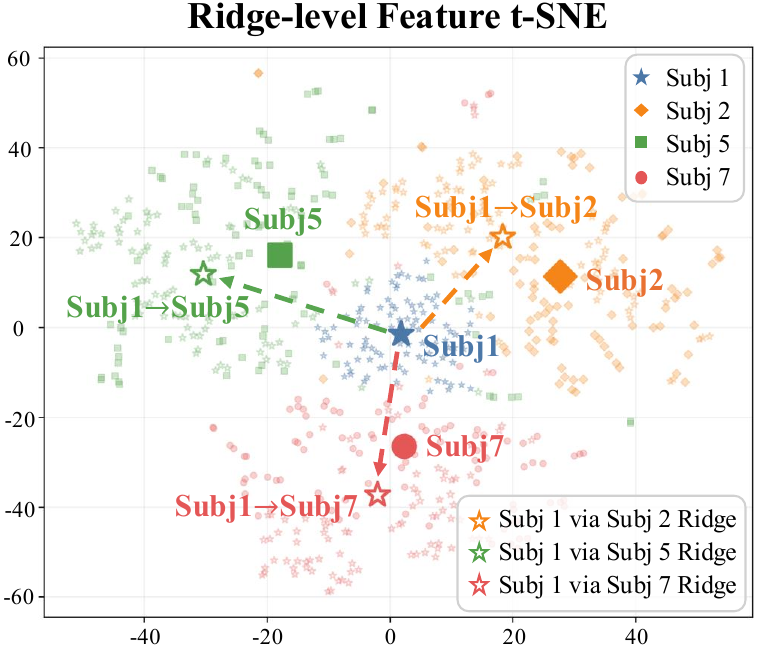}
    \caption{
  Ridge-level feature distributions visualized by t-SNE on Subject~1. We apply ridge mappings learned from different subjects to the same target-subject input from Subject~1 and visualize the resulting latent features. 
  The same target input is projected to different regions when using different subject-specific ridges, indicating that the ridge layer captures subject-dependent patterns. 
  % This observation motivates CSRR, which reuses source-subject ridge mappings as weak priors to model cross-subject projection variations and regularize target-subject adaptation under limited data.
}
  \label{fig: ridge}
\end{figure}

Existing cross-subject decoding methods typically introduce a subject-specific ridge projection to map voxel responses into a shared latent space, due to the mismatch in voxel dimensionality across subjects \cite{gong2025mindtuner,wang2024mindbridge,scotti2024mindeye2,dai2025mindaligner}. 
However, when only limited target-subject data are available during adaptation, the ridge projection learned for the new subject can be under-constrained and prone to overfitting, leading to unstable voxel-to-latent mapping. 
To investigate the role of ridge projections, we visualize in Fig.~\ref{fig: ridge} the latent features obtained by applying different subject-specific ridge mappings to the same target-subject input. 
Although the input fMRI responses are fixed, different ridge mappings project them to noticeably different regions in the latent space. 
This observation suggests that the ridge layer encodes subject-dependent projection patterns and can be used to model cross-subject variations in the voxel-to-latent mapping.
Based on this observation, we propose \textit{Cross-Subject Ridge Reuse} (CSRR), which incorporates source-subject ridge mappings as weak priors to regularize the target-subject projection during adaptation. 
Instead of treating the target ridge as an isolated mapping learned from limited data, CSRR exposes target adaptation to source-derived projection priors, thereby regularizing the target projection with cross-subject priors.

Let $V_t \in \mathbb{R}^{d_t}$ denote the voxel response of the target subject, and let the corresponding ridge mapper be parameterized by $(W_t,b_t)$. Its output is formulated as:
\begin{equation}
R_t = W_t V_t + b_t.
\end{equation}

Since source and target subjects may have different voxel dimensionalities, directly applying a source-subject ridge to target-subject responses requires a dimension-matching operation. We use a simple padding/truncation operator $\mathcal{A}(\cdot,d_s)$ to match the input dimensionality of the source ridge. This operation is not intended to establish voxel-wise anatomical correspondence across subjects. Instead, it enables each frozen source ridge to produce a weak output-space projection prior. The source-prior signal is therefore used only after projection, where multiple source ridges are aggregated to provide a consensus regularizer for the target ridge. 
For each source subject $s \in \mathcal{S}$ with voxel dimension $d_s$, we compute its projected feature under the ridge mapping of subject $s$:
\begin{equation}
R_s = W_s \mathcal{A}(V_t,d_s) + b_s.
\end{equation}
We aggregate the source-subject ridge outputs as
\begin{equation}
R_{\mathrm{src}} = \frac{1}{|\mathcal{S}|}\sum_{s \in \mathcal{S}} R_s.
\end{equation}
To inject source-subject priors into target-subject adaptation, we fuse the target-subject ridge output with the aggregated source prior:
\begin{equation}
R = (1-\alpha)R_t + \alpha R_{\mathrm{src}},
\label{eq: prior alpha}
\end{equation}
where $\alpha \in [0,1]$ controls the strength of prior injection.
This formulation balances the target-specific estimate $R_t$, which is learned from limited data, with the aggregated source prior $R_{\mathrm{src}}$, which provides a more stable reference induced by multiple source-subject projections. In this way, CSRR regularizes the under-constrained target ridge in the output space \cite{hastie2009elements}.
To further regularize the target-subject ridge without suppressing meaningful subject-specific adaptation, we introduce a weak cosine distillation term as below:
\begin{equation}
\mathcal{L}_{\mathrm{src}} = 1 - \cos\!\big(R_t,\operatorname{sg}(R_{\mathrm{src}})\big),
\end{equation}
where $\operatorname{sg}(\cdot)$ denotes stop-gradient. 
We introduce $w_{\mathrm{src}}$ as the weight of $\mathcal{L}_{\mathrm{src}}$ in the training loss. 
The cosine distillation term provides a weak constraint on the target ridge by aligning it with the aggregated source prior, while the stop-gradient operation and small loss weight prevent over-regularization. 
The final ridge representation $R$ is then fed into the downstream brain encoder. 
Overall, CSRR can be viewed as a lightweight output-space regularization mechanism \cite{scholkopf2002learning}, where multiple source ridges provide a consensus prior for stabilizing limited-data adaptation, consistent with multi-source adaptation principles \cite{ben2007analysis,mansour2009domain}. 
Importantly, CSRR does not assume exact voxel-wise correspondence across subjects. The source ridges are used only to generate weak projected priors after dimensionality matching and aggregation, rather than to directly decode target-subject voxels.

\subsection{Feature-Level Brain Augmentation}
Existing cross-subject decoding methods mainly emphasize latent alignment or lightweight adaptation of a shared decoder \cite{scotti2024mindeye2,gong2025mindtuner,wang2024mindbridge}. Although effective, such methods do not fully address the instability of brain-side representations during target-subject fine-tuning with limited data. In this setting, the model needs to preserve meaningful subject-specific characteristics, but may also become biased toward sample-specific patterns, which can weaken the shared cross-subject structure learned during pretraining. 
This limitation suggests the need to explicitly regularize intermediate brain representations during adaptation, rather than relying solely on latent alignment.
Motivated by prior methods \cite{guo2023aloft} suggesting that spectral domain provides a useful space for regularizing the model to learn robust semantic features, we empirically investigate cross-subject brain features in frequency domain. 
As illustrated in Fig.~\ref{fig: motivation}, the brain features extracted by the model exhibit noticeable cross-subject differences in their spectral statistics, with larger cross-subject discrepancy observed in amplitude-related statistics. 
Based on this observation, we propose \textit{feature-level brain augmentation}, a Fourier-based regularization strategy for limited-data target-subject adaptation.

Specifically, we perturb amplitude-related statistics while keeping the phase term unchanged, introducing controlled variation without disrupting the overall representation structure.
Given the $i$-th brain intermediate feature $x_i \in \mathbb{R}^{K}$, where $K$ denotes the latent space dimension, we first obtain its discrete Fourier transform \cite{nussbaumer1981fast,sundararajan2001discrete,jena2025discrete}:
{\setlength\abovedisplayskip{2pt}
\setlength\belowdisplayskip{2pt}
\begin{equation}
\mathcal{F}(x_i)^k = \sum_{n=0}^{K-1} x_i^n e^{-j 2\pi \frac{n}{K}k}, 
\end{equation}}
and its corresponding amplitude and phase are computed:
{\setlength\abovedisplayskip{2pt}
\setlength\belowdisplayskip{2pt}
\begin{equation}
\mathcal{A}(x_i) = \left[ R^2(x_i) + I^2(x_i) \right]^{\frac{1}{2}}, \mathcal{P}(x_i) = \arctan\!\left(\frac{I(x_i)}{R(x_i)}\right),
\end{equation}}
where $R(x_i)$ and $I(x_i)$ represent the real and imaginary part of $\mathcal{F}(x_i)$, respectively. 
We use Gaussian amplitude-statistics perturbation as a practical regularizer. Specifically, we compute the sample-level mean and standard deviation of the amplitude spectrum:
\begin{equation}
\mu(\mathcal{A}_i) = \frac{1}{K} \sum_{k=0}^{K-1} \mathcal{A}(x_i)^k,
\end{equation}
\begin{equation}
\sigma(\mathcal{A}_i)^2 = \frac{1}{K} \sum_{k=0}^{K-1} [\mathcal{A}(x_i)^k - \mu(\mathcal{A}(x_i))]^2.
\end{equation}

Inspired by previous advanced methods \cite{aguirre1997empirical,woolrich2025statistical,yao2025catd,wu2025bridging}, within a mini-batch, we estimate the batch-level variation of these statistics, \textit{i.e.}, the standard deviations of the statistics $\mu(\mathcal{A}_i)$ and $\sigma(\mathcal{A}_i)$, as follows:

{\setlength\abovedisplayskip{2pt}
\setlength\belowdisplayskip{2pt}
\begin{equation}
    \Sigma_{\mu}^2 = \frac{1}{B} \sum_{i}^{B} [\mu(\mathcal{A}(x_i)) - \mathbb{E}[\mu(\mathcal{A}(x_i))]]^2, 
\end{equation}}
{\setlength\abovedisplayskip{2pt}
\setlength\belowdisplayskip{2pt}
\begin{equation}
    \Sigma_{\sigma}^2 = \frac{1}{B} \sum_{i}^{B} [\sigma(\mathcal{A}(x_i)) - \mathbb{E}[\sigma(\mathcal{A}(x_i))]]^2.
\end{equation}}
To perturb amplitude statistics around their empirical mini-batch distribution while avoiding arbitrary feature corruption, we adopt Gaussian resampling to provide smooth and bounded statistical shifts, \textit{i.e.}, sampling perturbed statistics $\tilde{\mu}$ and $\tilde{\sigma}$ as:
\begin{equation}
 \tilde{\mu}(\mathcal{A}_i) \sim \mathcal{N}(\mu(\mathcal{A}_i), \Sigma_{\mu}^2), \;\; \tilde{\sigma}(\mathcal{A}_i) \sim \mathcal{N}(\sigma(\mathcal{A}_i), \Sigma_{\sigma}^2), 
\end{equation}
and reconstruct the perturbed amplitude as:
\begin{equation}
\tilde{\mathcal{A}}(x_i) = \tilde{\sigma}(\mathcal{A}_i) 
 \frac{\mathcal{A}(x_i) - \mu(\mathcal{A}_i)}{\sigma(\mathcal{A}_i)+\epsilon} + \tilde{\mu}(\mathcal{A}_i).
\end{equation}
The above resampled amplitude $\tilde{\mathcal{A}}(x_i)$ and the original phase $\mathcal{P}(x_i)$ are combined to form the augmented frequency: $\tilde{\mathcal{F}}(x_i) = \tilde{\mathcal{A}}(x_i) e^{j\mathcal{P}(x_i)}$, which is mapped back to the spatial domain to get the perturbed fMRI embeddings:
{\setlength\abovedisplayskip{2pt}
\setlength\belowdisplayskip{2pt}
\begin{equation}
 \tilde{x}_i^k
= \mathcal{F}^{-1}(\tilde{\mathcal{F}}(x_i))^k 
= \frac{1}{K}\sum_{n=0}^{K-1} \tilde{\mathcal{F}}(x_i)^n e^{j 2\pi \frac{n}{K}k},
\end{equation}}

By introducing controlled perturbations to spectral statistics, the proposed augmentation discourages the model from relying on unstable sample-specific patterns, as such patterns become inconsistent across augmented views and are less likely to be reinforced during training. 
By keeping the phase term unchanged and perturbing only amplitude statistics, the augmentation avoids excessive disruption to the original feature structure while exposing the model to controlled spectral variations.
We analyze the effect of different perturbations in Sec.~\ref{sec. gaussian modeling}, where Gaussian modeling yields diverse yet structured feature variations and leads to the best performance.

\subsection{Difficulty-Aware Image Blur}
Most existing cross-subject decoding methods directly use clean images or their pretrained visual embeddings as supervision targets \cite{scotti2024mindeye2,wang2024mindbridge,gong2025mindtuner}. However, under limited-data target-subject adaptation, such full-detail supervision can aggravate overfitting, since fine-grained image details (\textit{e.g.}, textures and background structures) are not always reliably reflected in brain responses.
To address this issue, we propose \textit{difficulty-aware image blur}, which adaptively controls fine-grained visual detail preserved in image-side supervision according to the current brain-image alignment. 
Specifically, samples that are easier to align may further overfit to fine-grained visual details under limited data, and therefore receive stronger blur regularization. In contrast, harder samples retain more visual structure, since excessive blur may further weaken their limited alignment signal.

\textit{(1) Difficulty estimation.}
Let $Z_B, Z_V \in \mathbb{R}^{B \times C}$ denote the normalized brain and clean-image embeddings in a mini-batch, respectively. Following the clean-image alignment setting, we first compute the sample-wise cosine similarity:
\begin{equation}
s_i = \cos(z_{b,i}, z_{x,i}),
\end{equation}
where $z_{b,i}$ and $z_{x,i}$ are the $i$-th rows of $Z_B$ and $Z_V$. We then standardize the similarity within the mini-batch and convert it into an easiness score:
\begin{equation}
\hat{s}_i = \frac{s_i - \mu_{\mathrm{batch}}}{\sigma_{\mathrm{batch}} + \epsilon},
\qquad
e_i = \sigma\!\left(\frac{\hat{s}_i}{T}\right),
\label{eq: easiness score}
\end{equation}
where $\sigma(\cdot)$ is the sigmoid function, $T$ is a temperature parameter, and $\epsilon$ is a small constant for numerical stability. To obtain a more stable estimate across training steps, we maintain an image-wise difficulty bank and update it using exponential moving average:
\begin{equation}
b_i \leftarrow m\, b_i + (1-m)\, e_i,
\label{eq: bank m}
\end{equation}
where $m \in [0,1)$ is the momentum coefficient. The corresponding hardness score is then defined as
\begin{equation}
h_i = 1 - b_i.
\end{equation}
A larger $h_i$ indicates that the sample is harder to align and should retain more unblurred visual structure during image-side supervision.

\textit{(2) Difficulty-aware blur construction.}
Given an image $I_i \in \mathbb{R}^{H \times W \times 3}$, we first generate a uniformly blurred version $I_i^{\mathrm{base}}$ using a Gaussian kernel \cite{wu2025bridging}:
\begin{equation}
I_i^{\mathrm{base}}(u,v)
=
\sum_{p=-k}^{k}\sum_{q=-k}^{k}
I_i(u-p,v-q)\,G(p,q),
\end{equation}
\begin{equation}
G(p,q)=\frac{1}{2\pi\sigma^2}\exp\!\left(-\frac{p^2+q^2}{2\sigma^2}\right).
\end{equation}
We then construct a spatial blending mask centered at the image center, where the clear-region radius is adjusted according to the sample difficulty:
\begin{equation}
r_i
=
\operatorname{clamp}\!\big(
\rho_i s (1 + \beta_h h_i),
\; r_{\min},\;
r_{\max}
\big),
\end{equation}
where $\operatorname{clamp}(x,a,b)=\min(\max(x,a),b)$, $s$ is a global scaling factor, and $\beta_h$ control hardness-aware expansion. 
$\rho_i$ denotes the LayerCAM-derived image-wise base radius ratio, where the salient region radius is estimated from the heatmap statistics and normalized by the image size \cite{jiang2021layercam}.
Since $r_i$ increases with the hardness score $h_i$, harder samples preserve a larger clear region and thus retain more semantic and structural cues. Easier samples receive a smaller clear region and stronger peripheral blur, which reduces the risk of overfitting to unreliable fine-grained details.
The blending mask is:
% Based on the adaptive radius, we define a blending mask
\begin{equation}
\alpha_i(u,v)
=
\exp\!\left(
-\lambda_{\alpha}
\sqrt{
\frac{(u-c_h)^2+(v-c_w)^2}{r_i^2+\epsilon}
}
\right),
\label{eq: blending mask}
\end{equation}
where $\lambda_{\alpha}$ controls the decay rate from the center to the periphery \cite{wu2025bridging}. The final blurred image is constructed as
\begin{equation}
I_i^{\mathrm{blur}}
=
\alpha_i \odot I_i + (1-\alpha_i)\odot I_i^{\mathrm{base}},
\end{equation}
where $\odot$ denotes element-wise multiplication. The blurred-image embedding is obtained by $Z_V^{\mathrm{blur}} = F_V(I^{\mathrm{blur}})$.

\textit{(3) Difficulty-aware auxiliary supervision.}
We use the blurred image as an auxiliary supervision target together with the clean image. 
The blurred-image alignment loss is $\mathcal{L}_{\mathrm{clip}}^{\mathrm{blur}} = \mathcal{L}_{\mathrm{clip}}(Z_B,Z_V^{\mathrm{blur}})$, and the image-side alignment objective is
\begin{equation}
\mathcal{L}_{\mathrm{clip}}^{\mathrm{all}}
=
\mathcal{L}_{\mathrm{clip}}(Z_B,Z_V)
+
w_{\mathrm{blur}}\,\mathcal{L}_{\mathrm{clip}}(Z_B,Z_V^{\mathrm{blur}}).
\end{equation}

This module is designed to improve the reliability of image-side supervision under limited-data adaptation. By assigning stronger blur regularization to easy samples and adopting a more conservative strategy for hard samples, the proposed method downweights fine-grained visual details that may be less reliably reflected in brain responses, while preserving sufficient visual structure for reliable cross-modal alignment.

\section{Experiments and Results}
\label{sec:formatting}

\subsection{Datasets and Implementation Details}
\textbf{Datasets.} We conduct experiments on the Natural Scenes Dataset (NSD) \cite{allen2022massive}, the largest public $7$T fMRI dataset containing brain responses from $8$ subjects viewing natural images from MS-COCO \cite{lin2014microsoft}. 
Each subject completed $30$-$40$ one-hour sessions (approximately $750$ trials per session, with $3$ seconds per image). 
Following recent cross-subject decoding works under limited-data adaptation \cite{scotti2024mindeye2,gong2025mindtuner,dai2025mindaligner}, we use only a single fMRI session (containing one hour of fMRI data) from the target subject during adaptation for method evaluation.
Following \cite{scotti2024mindeye2,gong2025mindtuner}, we pretrain the shared model on data from $7$ subjects and fine-tune it on a held-out target subject. 
For all quantitative experiments, we use subjects ${1,2,5,7}$ as the target-subject evaluation set, as they are the four NSD subjects who completed all 40 sessions. The corresponding voxel counts are $15{,}724$, $14{,}278$, $13{,}039$, and $12{,}682$, respectively. Unless otherwise specified, all results are averaged over these four subjects under a unified 1-hour adaptation protocol.

\textbf{Implementation Details.}
For each subject, we train a subject-specific ridge layer to project the input voxel response into a shared $4{,}096$-dimensional latent space. The downstream decoder is first pretrained on the source subjects and then fine-tuned on the held-out target subject. All fine-tuning experiments are conducted for $150$ epochs on a single Tesla A800 80GB GPU, using a batch size of $10$ and the AdamW optimizer with a learning rate of $3\times10^{-4}$.
The weights of the low-level reconstruction loss, diffusion-prior loss, and skip-LoRA correlation loss are set to $0.5$, $1.0$, and $1.5$, respectively. The ranks of all LoRA and Skip-LoRA modules are set to $8$. 
During fine-tuning, we use BiMixCo loss in the first one-third of training and SoftCLIP loss in the remaining epochs. The final reconstructed image is obtained by weighted averaging of the high-level reconstruction and the low-level blurry reconstruction with a ratio of $3{:}1$.
For StableMind, the weight of the blur-based alignment loss $\mathcal{L}_{\mathrm{clip}}^{\mathrm{blur}}$ is set to $0.50$. The Gaussian blur kernel size is set to $51$.
The source-prior fusion weight $\alpha$ in Eq.~(\ref{eq: prior alpha}) is set to $0.1$, the momentum $m$ used to update the image-wise difficulty bank in Eq.~(\ref{eq: bank m}) is set to $0.85$, the temperature $T$ in Eq.~(\ref{eq: easiness score}) is set to $0.028$, the global radius scaling factor $s$ is set to $0.92$, $\beta_h$ is $0.18$, and $\lambda_\alpha$ in Eq.~(\ref{eq: blending mask}) is $3$.
$r_{min}$ and $r_{max}$ are $0.18$ and $0.28$.
% \needrev{$s$ is the global radius scaling factor, $\beta_h$ and $\beta_e$ control hardness-aware expansion and easiness-aware shrinkage, respectively, and $\lambda_\alpha$ controls the spatial decay of the blending mask in Eq.~(22).}
For retrieval evaluation, we follow \cite{scotti2024mindeye2} by computing Top-1 accuracy over a candidate pool of $300$ samples (chance level $1/300$), where cosine similarity is used for matching, and the results are averaged over $30$ random trials to reduce sampling variance.
Unless otherwise specified, the reported ``4.04M'' trainable parameter count includes all trainable adaptation modules excluding subject-specific ridge layers, following the counting protocol used in prior comparisons. The frozen source-subject ridge layers used by CSRR are retained as part of the pretrained decoder and are not updated during target adaptation. We therefore report them separately from trainable adaptation parameters.

\begin{table*}[!t]\normalsize
\centering
\caption{
Comparison with state-of-the-art methods on NSD under the unified 1-hour adaptation protocol. \\We report the four-subject average and the per-subject results on Subjects $1$, $2$, $5$, and $7$. 
We report low-level (PixCorr, SSIM, Alex(2/5)), high-level (Inception, CLIP, EffNet, SwAV), and retrieval (Image/Brain) metrics.
``Tr. Param.'' counts trainable adaptation parameters. Best is \textbf{Bold}.
}
\resizebox{\textwidth}{!}{
\begin{tabular}{l|cccc|cccc|cc}
\toprule
\multirow{2}{*}{\textbf{Method}} &
\multicolumn{4}{c|}{\textbf{Low-Level}} & 
\multicolumn{4}{c|}{\textbf{High-Level}} & 
\multicolumn{2}{c}{\textbf{Retrieval}} \\ \cline{2-11} \noalign{\vskip 1.5pt} 
 & PixCorr$\uparrow$ & SSIM$\uparrow$ & Alex(2)$\uparrow$ & Alex(5)$\uparrow$ & Incep$\uparrow$ & CLIP$\uparrow$ & Eff.$\downarrow$ & SwAV$\downarrow$  & Image$\uparrow$ & Brain$\uparrow$ \\ 
 \midrule
MindEye2 (1h) \cite{scotti2024mindeye2} {\scriptsize (ICML'24)} & 0.195 & 0.419 & 84.21\% & 90.54\% & 80.66\% & 80.00\% & 0.805 & 0.465 & 78.97\% & 57.39\% \\
MindBridge (1h) \cite{wang2024mindbridge} {\scriptsize (CVPR'24)} & 0.112 & 0.229 & 79.60\% & 89.00\% & 82.30\% & \textbf{86.70\%} & 0.840 & 0.521 & 79.60\% & 56.70\% \\
MindAligner (1h) \cite{dai2025mindaligner} {\scriptsize (ICML'25)} & 0.206 & 0.414 & 85.60\% & 91.63\% & 82.13\% & 81.08\% & 0.802 & 0.463 & 78.93\% & 75.26\% \\
MindTuner (1h) \cite{gong2025mindtuner} {\scriptsize (AAAI'25)} & 0.224 & \textbf{0.420} & 87.70\% & 93.63\% & 84.75\% & 83.45\% & \textbf{0.781} & \textbf{0.440} & 83.00\% & 75.95\% \\
\rowcolor{myblue} \textbf{StableMind} (1h) {\scriptsize (Ours)} & \textbf{0.226} & 0.415 & \textbf{87.83\%} & \textbf{93.63\%} & \textbf{85.08\%} & 83.56\% & 0.784 & 0.445 & \textbf{84.02\%} & \textbf{81.66\%} \\
\midrule
MindEye2 (subj1) \cite{scotti2024mindeye2} {\scriptsize (ICML'24)} & 0.235 & \textbf{0.428} & 88.02\% & 93.33\% & 83.56\% & 81.76\% & 0.798 & 0.459 & 93.96\% & 77.63\% \\
MindAligner (subj1) \cite{dai2025mindaligner} {\scriptsize (ICML'25)} & 0.226 & 0.415 & 88.19\% & 93.26\% & 83.48\% & 81.76\% & 0.800 & 0.459 & 90.90\% & 86.88\% \\
MindTuner (subj1) \cite{gong2025mindtuner} {\scriptsize (AAAI'25)} & \textbf{0.262} & 0.422 & 90.60\% & 94.90\% & 85.80\% & 84.60\% & 0.774 & 0.433 & 94.20\% & 87.40\% \\
\rowcolor{myblue} \textbf{StableMind} (subj1) {\scriptsize (Ours)} & 0.255 & 0.418 & \textbf{90.88\%} & \textbf{95.41\%} & \textbf{86.42\%} & \textbf{85.31\%} & \textbf{0.768} & \textbf{0.433} & \textbf{95.07\%} & \textbf{92.96\%} \\
\midrule
MindEye2 (subj2) \cite{scotti2024mindeye2} {\scriptsize (ICML'24)} & 0.200 & \textbf{0.433} & 85.00\% & 92.13\% & 81.86\% & 79.89\% & 0.807 & 0.454 & 90.53\% & 67.18\% \\
MindAligner (subj2) \cite{dai2025mindaligner} {\scriptsize (ICML'25)} & 0.218 & 0.426 & 88.08\% & 93.33\% & 82.47\% & 81.62\% & 0.791 & 0.452 & 90.04\% & 85.61\% \\
MindTuner (subj2) \cite{gong2025mindtuner} {\scriptsize (AAAI'25)} & 0.225 & 0.425 & 89.10\% & \textbf{95.10\%} & 84.80\% & 83.70\% & 0.781 & \textbf{0.440} & 93.70\% & 82.80\% \\
\rowcolor{myblue} \textbf{StableMind} (subj2) {\scriptsize (Ours)} & \textbf{0.232} & 0.426 & \textbf{89.30\%} & 94.96\% & \textbf{86.11\%} & \textbf{84.41\%} & \textbf{0.779} & 0.441 & \textbf{94.04\%} & \textbf{92.01\%} \\
\midrule
MindEye2 (subj5) \cite{scotti2024mindeye2} {\scriptsize (ICML'24)} & 0.175 & 0.405 & 83.11\% & 90.83\% & 82.32\% & 78.53\% & 0.781 & 0.444 & 66.94\% & 46.96\% \\
MindAligner (subj5) \cite{dai2025mindaligner} {\scriptsize (ICML'25)} & 0.197 & 0.409 & 84.69\% & 91.61\% & 82.63\% & 80.12\% & 0.784 & 0.454 & 70.62\% & 65.95\% \\
MindTuner (subj5) \cite{gong2025mindtuner} {\scriptsize (AAAI'25)} & 0.208 & \textbf{0.415} & \textbf{86.60\%} & \textbf{93.70\%} & \textbf{87.70\%} & \textbf{85.90\%} & \textbf{0.750} & \textbf{0.422} & 72.20\% & 68.10\% \\
\rowcolor{myblue} \textbf{StableMind} (subj5) {\scriptsize (Ours)} & \textbf{0.211} & 0.413 & 86.49\% & 93.10\% & 86.00\% & 84.28\% & 0.773 & 0.439 & \textbf{75.04\%} & \textbf{71.14\%} \\
\midrule
MindEye2 (subj7) \cite{scotti2024mindeye2} {\scriptsize (ICML'24)} & 0.170 & 0.408 & 80.70\% & 85.90\% & 74.90\% & 79.84\% & 0.834 & 0.504 & 64.44\% & 37.77\% \\
MindAligner (subj7) \cite{dai2025mindaligner} {\scriptsize (ICML'25)} & 0.183 & 0.407 & 81.45\% & 88.31\% & 79.92\% & \textbf{80.83\%} & 0.834 & 0.487 & 64.18\% & 62.58\% \\
MindTuner (subj7) \cite{gong2025mindtuner} {\scriptsize (AAAI'25)} & 0.202 & \textbf{0.417} & 84.50\% & 90.80\% & 80.70\% & 79.60\% & 0.817 & \textbf{0.465} & 71.90\% & 65.50\% \\
\rowcolor{myblue} \textbf{StableMind} (subj7) {\scriptsize (Ours)} & \textbf{0.206} & 0.403 & \textbf{84.66\%} & \textbf{91.10\%} & \textbf{81.77\%} & 80.24\% & \textbf{0.817} & 0.467 & \textbf{71.94\%} & \textbf{70.53\%}\\
\bottomrule
\end{tabular}}
        % \vspace{-0.4cm}
\label{tab: SOTA comparison draft}
\end{table*}

\begin{table}[!t]\normalsize
\centering
\caption{
Efficiency comparison results. “Tr. Param.” refers to the
model's trainable parameters when adding a new subject.
Ridge regression parameters correspond to Subject 1.
}
\resizebox{\linewidth}{!}{
\begin{tabular}{l|ccc|c}
\toprule
\multirow{2}{*}{\textbf{Method}} & \multicolumn{3}{c|}{\textbf{Training Parameters}}  & \multirow{2}{*}{\textbf{Inference}} \\
& Ridge & Backbone & Adapter  &  \\
\midrule
MindEye2 \cite{scotti2024mindeye2} & 64.41M & 1903M & 260M & 9.447s \\
MindAligner \cite{dai2025mindaligner} & 139.23M & 0M & 0M & 9.573s \\
MindTuner \cite{gong2025mindtuner} & 64.41M & 12.30M & 0M & - \\
\rowcolor{myblue} \textbf{StableMind} {\scriptsize (Ours)} & 64.41M & 4.04M & 0M & 9.654s \\
\bottomrule
\end{tabular}
}
\label{tab: efficiency}
\end{table}

\subsection{Comparison with SOTA methods}

\textbf{Image and Brain Retrievals.}
Tab.~\ref{tab: SOTA comparison draft} illustrates the retrieval comparison with recent cross-subject brain decoding methods under the unified 1-hour adaptation protocol, including the image retrieval that retrieves image embeddings with the highest cosine similarity based on fMRI embeddings, and the brain retrieval that retrieves fMRI embeddings using the highest similarity with image embeddings.
On the four-subject average, StableMind achieves $84.02\%$ image retrieval accuracy and $81.66\%$ brain retrieval accuracy, improving over the SOTA method MindTuner \cite{gong2025mindtuner} by $1.02\%$ ($84.02\%$ vs. $83.00\%$) and $5.71\%$ ($81.66\%$ vs. $75.95\%$), respectively, and substantially outperforming MindEye2 \cite{scotti2024mindeye2} on brain retrieval by $24.27\%$($81.66\%$ vs. $57.39\%$). 
These improvements indicate that StableMind yields more discriminative brain-image representations under limited-data subject transfer.
The combination of source-guided projection regularization, feature-level Fourier augmentation, and difficulty-aware image-side supervision helps stabilize the alignment between brain and image representations while reducing the effect of subject-specific variations. 
The per-subject results show that StableMind achieves the best retrieval performance on all four target subjects. 
These results indicate that StableMind provides effective cross-modal representation alignment and improves cross-subject adaptation under the limited 1-hour setting.

\textbf{Brain-to-Image Reconstruction.}
We evaluate fMRI-to-image reconstruction quality using both low-level metrics (PixCorr, SSIM \cite{wang2004image}, AlexNet(2), and AlexNet(5) \cite{krizhevsky2012imagenet}) and high-level metrics (Inception \cite{szegedy2016rethinking}, CLIP \cite{radford2021learning}, EffNet \cite{tan2019efficientnet}, and SwAV \cite{caron2020unsupervised}). 
As shown in Tab.~\ref{tab: SOTA comparison draft}, averaged over the four target subjects, StableMind improves over MindEye2 \cite{scotti2024mindeye2} on most reconstruction metrics, increasing PixCorr from $0.195$ to $0.226$, Alex(2) from $84.21\%$ to $87.83\%$, Inception from $80.66\%$ to $85.08\%$, and CLIP from $80.00\%$ to $83.56\%$. 
Compared with MindTuner\cite{gong2025mindtuner}, StableMind achieves broadly comparable reconstruction performance, with advantages on several high-level metrics and slight degradation on others, while using substantially fewer trainable adaptation parameters($4.04\mathrm{M}$ vs. $12.30\mathrm{M}$). 
This result suggests that StableMind provides a parameter-efficient adaptation strategy while maintaining competitive reconstruction quality.
Overall, these results suggest that StableMind provides a competitive and parameter-efficient solution for cross-subject fMRI-to-image decoding, achieving clear retrieval gains and broadly comparable reconstruction quality under the same protocol.

\subsection{Ablation Studies}

\begin{table*}[!t]
\centering
\caption{
Ablation of the three main components on Subject $1$ under the 1-hour adaptation protocol. CSRR, FBA, and DIB denote cross-subject ridge reuse, feature-level brain augmentation, and difficulty-aware image blur, respectively.
}
\resizebox{\textwidth}{!}{
\begin{tabular}{ccc|cccc|cccc|cc}
\toprule
\multicolumn{3}{c|}{\textbf{Method}} &
\multicolumn{4}{c|}{\textbf{Low-Level}} & 
\multicolumn{4}{c|}{\textbf{High-Level}} & 
\multicolumn{2}{c}{\textbf{Retrieval}} \\ 
\midrule
CSRR & FBA & DIB & PixCorr$\uparrow$ & SSIM$\uparrow$ & Alex(2)$\uparrow$ & Alex(5)$\uparrow$ & Incep$\uparrow$ & CLIP$\uparrow$ & Eff.$\downarrow$ & SwAV$\downarrow$  & Image$\uparrow$ & Brain$\uparrow$ \\ 
\midrule
- & - & - & 0.241 & 0.414 & 89.71\% & 94.54\% & 84.78\% & 82.90\% & 0.788 & 0.446 & 93.58\% & 89.94\% \\
- & \checkmark & \checkmark & 0.248 & 0.415 & 90.20\% & 95.04\% & 85.15\% & 83.94\% & 0.781 & 0.441 & 94.64\% & 92.61\% \\
\checkmark & - & \checkmark & 0.245 & 0.414 & 89.65\% & 94.50\% & 85.36\% & 84.28\% & 0.783 & 0.440 & 94.71\% & 92.31\% \\
\checkmark & \checkmark & -  & 0.252 & 0.416 & 90.37\% & 95.20\% & 85.17\% & 84.21\% & 0.771 & 0.437 & 94.13\% & 92.51\% \\
\midrule
\rowcolor{myblue} \checkmark & \checkmark & \checkmark & \textbf{0.255} & \textbf{0.418} & \textbf{90.88\%} & \textbf{95.41\%} & \textbf{86.42\%} & \textbf{85.31\%} & \textbf{0.768} & \textbf{0.433} & \textbf{95.07\%} & \textbf{92.96\%} \\
\bottomrule
\end{tabular}}
% }
\label{tab: ablation}
        % \vspace{-0.4cm}
\end{table*}

\begin{table*}[!t]
\centering
\caption{Ablation study of cross-subject ridge reuse (CSRR) on Subject~1. ``SourceFuse'' denotes feature fusion with source priors, while ``CosLoss'' is the cosine distillation loss.}
\resizebox{\textwidth}{!}{
  \setlength{\tabcolsep}{2.0mm}{
\begin{tabular}{cc|cccc|cccc|cc}
\toprule
\multicolumn{2}{c|}{\textbf{Variant}} & \multicolumn{4}{c|}{\textbf{Low-Level}} & 
\multicolumn{4}{c|}{\textbf{High-Level}} & 
\multicolumn{2}{c}{\textbf{Retrieval}} \\
\midrule
SourceFuse & CosLoss & PixCorr$\uparrow$ & SSIM$\uparrow$ & Alex(2)$\uparrow$ & Alex(5)$\uparrow$ & Incep$\uparrow$ & CLIP$\uparrow$ & Eff.$\downarrow$ & SwAV$\downarrow$  & Image$\uparrow$ & Brain$\uparrow$ \\ 
\midrule
- & - & 0.248 & 0.415 & 90.20\% & 95.04\% & 85.15\% & 83.94\% & 0.781 & 0.441 & 94.64\% & 92.61\% \\
- & \checkmark & 0.250 & 0.417 & 90.43\% & 95.16\% & 86.24\% & 84.79\% & 0.770 & 0.434 & 94.50\% & 91.93\% \\
\checkmark & - & 0.251 & 0.417 & 90.88\% & 95.32\% & 86.33\% & 84.62\% & 0.775 & 0.436 & 94.61\% & 92.07\% \\
\rowcolor{myblue} \checkmark & \checkmark & \textbf{0.255} & \textbf{0.418} & \textbf{90.88\%} & \textbf{95.41\%} & \textbf{86.42\%} & \textbf{85.31\%} & \textbf{0.768} & \textbf{0.433} & \textbf{95.07\%} & \textbf{92.96\%} \\
\bottomrule
\end{tabular}}
}
\label{tab: CSRR Ablation}
\end{table*}

\begin{figure}[tb!]
    % \centering
        \subfloat[Weight of $\mathcal{L}_{\mathrm{src}}$.]{
          \includegraphics[width=0.49\linewidth]{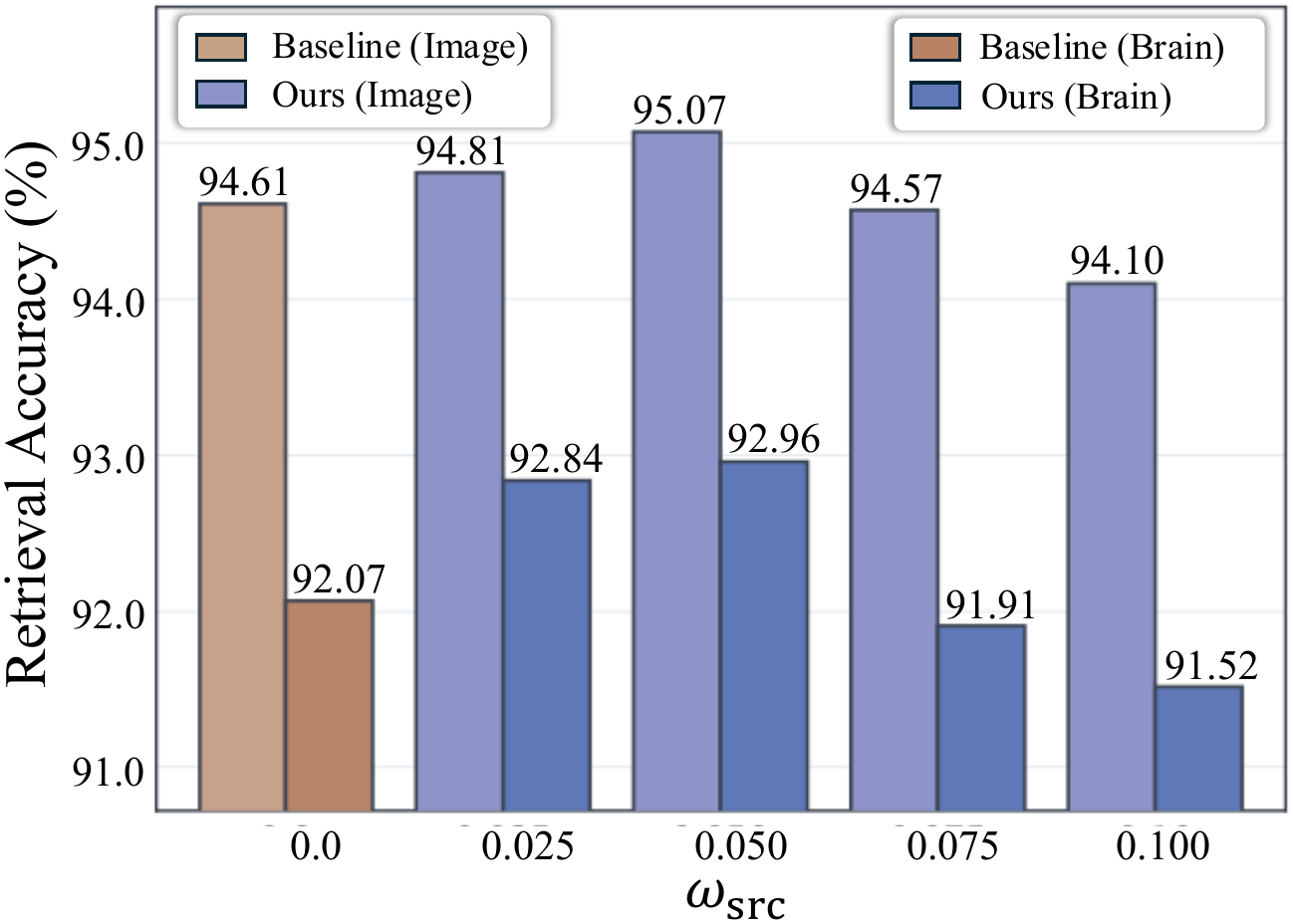}
          \label{fig:cons}
 }
        \subfloat[Weight of $\mathcal{L}_{\mathrm{clip}}^{\mathrm{blur}}$.]{
            \includegraphics[width=0.49\linewidth]{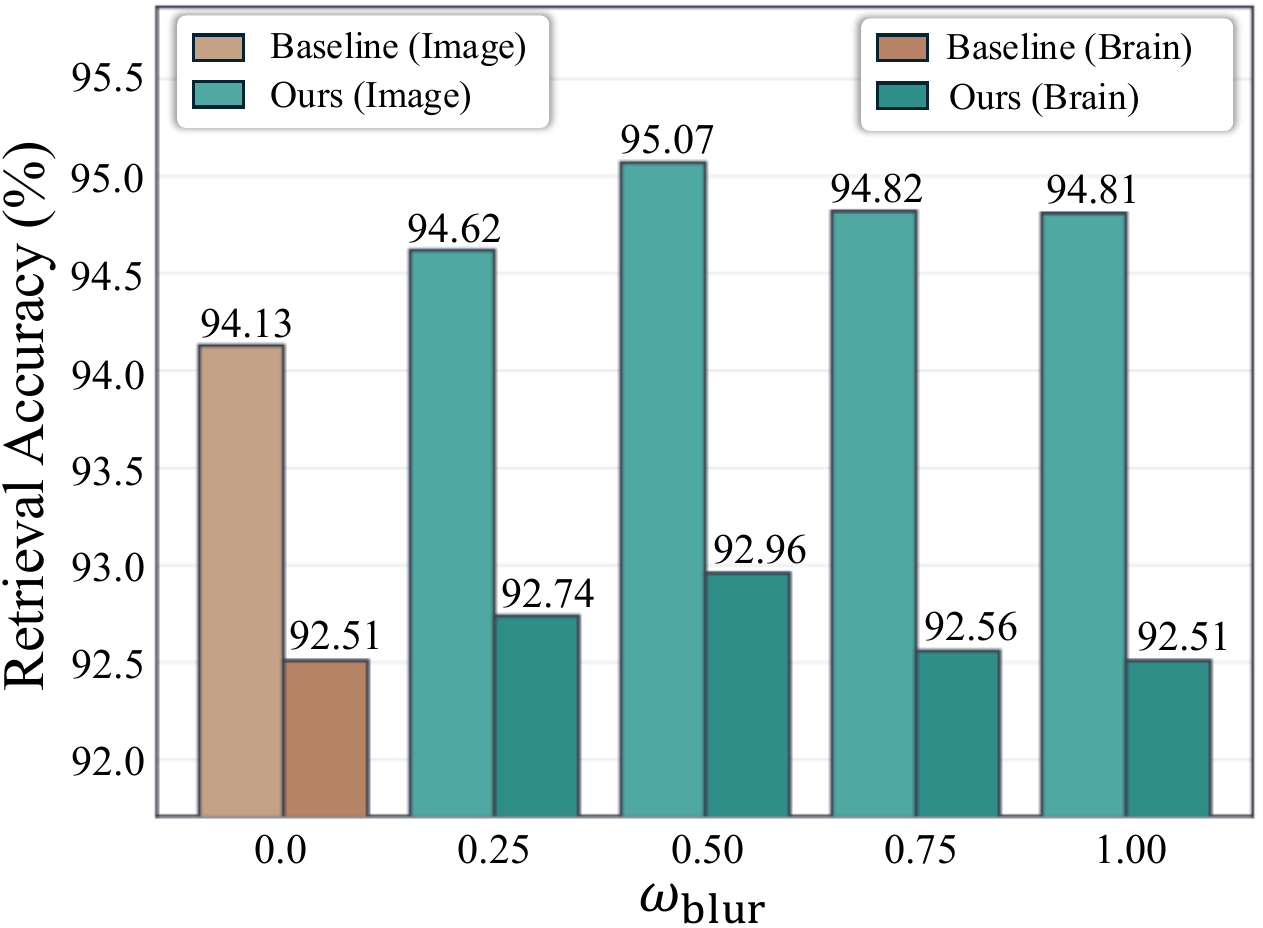}
            \label{fig:blur}
 }
        % \vspace{-0.3cm}
        \caption{          
          Parameter sensitivity of StableMind to the weight $w_{\mathrm{src}}$ of the source-prior consistency loss $\mathcal{L}_{\mathrm{src}}$ and the weight $w_{\mathrm{blur}}$ of the blurred-image alignment loss $\mathcal{L}_{\mathrm{clip}}^{\mathrm{blur}}$. 
          Experiments are conducted on Subject~1.
 }
        \label{fig: weight}
        % \vspace{-0.4cm}
\end{figure}

\textbf{Ablation Study of Main Components.} 
We further analyze the contribution of the three main components in StableMind, namely cross-subject ridge reuse (CSRR), feature-level brain augmentation (FBA), and difficulty-aware image blur (DIB). 
The experiments are conducted on Subject~1 under the 1-hour adaptation protocol. As shown in Tab.~\ref{tab: ablation}, all three two-module variants improve over the baseline without any additional module, indicating that each component contributes useful regularization under limited-data adaptation. In particular, the combination of FBA and DIB improves PixCorr from $0.241$ to $0.248$, increases image retrieval by $1.06\%$ ($94.64\%$ vs. $93.58\%$) and brain retrieval by $2.67\%$ ($92.61\%$ vs. $89.94\%$), suggesting that jointly regularizing brain-side features and image-side supervision already provides a clear benefit. The combination of CSRR and DIB improves Inception from $84.78\%$ to $85.36\%$ and CLIP from $82.90\%$ to $84.28\%$, while also improving retrieval performance, indicating that stabilizing the input projection and refining image-side supervision are both beneficial for semantic alignment. Similarly, combining CSRR and FBA yields consistent gains on low-level and high-level reconstruction metrics, increasing PixCorr to $0.252$ and Inception to $85.17\%$, which supports the role of brain-side regularization in improving adaptation robustness.
When all three components are enabled, StableMind achieves the best performance across all reported metrics. 
These results show that the three components are complementary, \textit{i.e.}, CSRR regularizes subject-level transfer, FBA improves robustness to subject-specific feature variation, and DIB further provides reliable image-side supervision during alignment.

\textbf{Parameter Sensitivity.}
We investigate the sensitivity of StableMind to two key hyperparameters, \textit{i.e.}, the weight $w_{\mathrm{src}}$ of the source-prior consistency loss $\mathcal{L}_{\mathrm{src}}$ and the weight $w_{\mathrm{blur}}$ of the blurred-image alignment loss $\mathcal{L}_{\mathrm{clip}}^{\mathrm{blur}}$.
As shown in Fig.~\ref{fig: weight}, we vary $w_{\mathrm{src}}$ in $\{0.0,0.025,0.050,0.075,0.100\}$ and $w_{\mathrm{blur}}$ in $\{0.0,0.25,0.50,0.75,1.00\}$, reporting both image retrieval and brain retrieval accuracies.
For $\mathcal{L}_{\mathrm{src}}$, enabling source-prior consistency improves both retrieval directions over the $0.0$ setting. 
The best performance is achieved at $w_{\mathrm{src}}=0.05$, where image retrieval increases from $94.61\%$ to $95.07\%$ and brain retrieval increases from $92.07\%$ to $92.96\%$.
Larger weights lead to a slight decline, suggesting that overly strong source-prior regularization may suppress target-specific adaptation.
For $\mathcal{L}_{\mathrm{clip}}^{\mathrm{blur}}$, blur-aware supervision also brings clear gains over the clean-only setting. 
The best trade-off is obtained at $w_{\mathrm{blur}}=0.50$, improving image retrieval from $94.13\%$ to $95.07\%$ and brain retrieval from $92.51\%$ to $92.96\%$. 
When $w_{\mathrm{blur}}$ is further increased, the performance gradually decreases but remains comparable to the baseline, indicating that moderate blur supervision is beneficial while excessive blur may weaken useful visual details.
Therefore, we set $w_{\mathrm{src}}=0.05$ and $w_{\mathrm{blur}}=0.50$ as the default values in all experiments.

\begin{table*}[!t]
\centering
\caption{
Comparison of different source-prior construction strategies in CSRR on Subject~1 under the unified 1-hour adaptation protocol. 
``Random One'' randomly selects a single source-subject ridge prior, ``Nearest One'' uses the most similar source subject, and ``All Average'' aggregates all available source priors. 
}
\resizebox{\textwidth}{!}{
  \setlength{\tabcolsep}{2.3mm}{
\begin{tabular}{l|cccc|cccc|cc}
\toprule
\multirow{2}{*}{\textbf{Source Prior}} & \multicolumn{4}{c|}{\textbf{Low-Level}} & 
\multicolumn{4}{c|}{\textbf{High-Level}} & 
\multicolumn{2}{c}{\textbf{Retrieval}} \\ \cline{2-11} \noalign{\vskip 1.5pt} 
 & PixCorr$\uparrow$ & SSIM$\uparrow$ & Alex(2)$\uparrow$ & Alex(5)$\uparrow$ & Incep$\uparrow$ & CLIP$\uparrow$ & Eff.$\downarrow$ & SwAV$\downarrow$  & Image$\uparrow$ & Brain$\uparrow$ \\ 
\midrule
Random One & 0.251 & 0.416 & 90.71\% & 95.17\% & 85.70\% & 84.25\% & 0.773 & 0.436 & 94.53\% & 92.25\% \\
Nearest One& 0.255 & 0.417 & 90.78\% &95.20\% & 85.76\% & 83.67\% & 0.776 & 0.438 & 94.42\% & 92.10\%\\
\rowcolor{myblue} All Average & \textbf{0.255} & \textbf{0.418} & \textbf{90.88\%} & \textbf{95.41\%} & \textbf{86.42\%} & \textbf{85.31\%} & \textbf{0.768} & \textbf{0.433} & \textbf{95.07\%} & \textbf{92.96\%} \\
\bottomrule
\end{tabular}}
}
\label{tab: source prior construct}
\vspace{0.1cm}
\end{table*}

\begin{table*}[!t]
\centering
\caption{
Comparison of different image-side supervision strategies on Subject~1 under the unified 1-hour adaptation protocol. 
``Clean Image'' uses the original image supervision, ``Whole blur'' applies uniform blur to the entire image, ``Fixed-radius blur'' uses a predefined blur radius, and ``Difficulty-aware blur'' adaptively adjusts blur based on alignment difficulty. 
}
\resizebox{\textwidth}{!}{
      \setlength{\tabcolsep}{2.mm}{
\begin{tabular}{l|cccc|cccc|cc}
\toprule
\multirow{2}{*}{\textbf{Image-Side Blur}} & \multicolumn{4}{c|}{\textbf{Low-Level}} & 
\multicolumn{4}{c|}{\textbf{High-Level}} & 
\multicolumn{2}{c}{\textbf{Retrieval}} \\ \cline{2-11} \noalign{\vskip 1.5pt} 
 & PixCorr$\uparrow$ & SSIM$\uparrow$ & Alex(2)$\uparrow$ & Alex(5)$\uparrow$ & Incep$\uparrow$ & CLIP$\uparrow$ & Eff.$\downarrow$ & SwAV$\downarrow$  & Image$\uparrow$ & Brain$\uparrow$ \\ 
\midrule
Clean image & 0.252 & 0.416 & 90.37\% & 95.20\% & 85.17\% & 83.21\% & 0.771 & 0.437 & 94.13\% & 92.51\% \\
Whole blur & 0.253 & 0.417 & 90.57\% & 95.03\% & 85.68\% & 84.26\% & 0.774 & 0.438 & 93.84\% & 92.18\% \\
Fixed-radius blur & 0.251 & 0.417 & 90.68\% & 95.18\% & 85.93\% & 84.70\% & 0.772 & 0.437 & 94.60\% & 92.69\% \\ 
\rowcolor{myblue} Difficulty-aware blur & \textbf{0.255} & \textbf{0.418} & \textbf{90.88\%} & \textbf{95.41\%} & \textbf{86.42\%} & \textbf{85.31\%} & \textbf{0.768} & \textbf{0.433} & \textbf{95.07\%} & \textbf{92.96\%} \\
\bottomrule
\end{tabular}}}
\label{tab: blur strategies}
\vspace{0.1cm}
\end{table*}

\begin{table*}[!t]
\centering
\caption{
Comparison of different feature-level brain augmentation (FBA) variants on Subject~1 under the unified 1-hour adaptation protocol. 
``No FBA'' removes feature-level augmentation, ``Random Noise'' adds unstructured random perturbation, ``Uniform'' samples amplitude statistics from a uniform distribution, ``Swap'' replaces amplitude statistics with those from another sample, ``Mix'' interpolates amplitude statistics between samples, and ``Gaussian'' is the Gaussian amplitude-statistics perturbation.
}
\resizebox{\textwidth}{!}{
    \setlength{\tabcolsep}{2.2mm}{
\begin{tabular}{l|cccc|cccc|cc}
\toprule
\multirow{2}{*}{\shortstack{\textbf{Feature}\\\textbf{Augmentation}}} 
% \multirow{2}{*}{\shortstack{\textbf{FBA}\\\textbf{Variant}}}
& \multicolumn{4}{c|}{\textbf{Low-Level}} & 
\multicolumn{4}{c|}{\textbf{High-Level}} & 
\multicolumn{2}{c}{\textbf{Retrieval}} \\ \cline{2-11} \noalign{\vskip 1.5pt} 
 & PixCorr$\uparrow$ & SSIM$\uparrow$ & Alex(2)$\uparrow$ & Alex(5)$\uparrow$ & Incep$\uparrow$ & CLIP$\uparrow$ & Eff.$\downarrow$ & SwAV$\downarrow$  & Image$\uparrow$ & Brain$\uparrow$ \\ 
\midrule
Vanilla & 0.245 & 0.414 & 89.65\% & 94.50\% & 85.36\% & 84.28\% & 0.783 & 0.440 & 94.71\% & 92.31\% \\
\midrule
Random Noise & 0.246 & 0.415 & 90.01\% & 94.80\%  & 84.99\% & 83.27\% & 0.781 & 0.443 & 94.09\% & 91.39\% \\
Uniform Model & 0.245 & 0.416 & 89.83\% & 94.61\% & 84.64\% & 84.44\% & 0.778 & 0.445 & 94.47\% & 91.08\% \\
Swap Amplitude & 0.250 & 0.414 & 90.34\% & 95.12\% & 86.17\% & 84.64\% & 0.775 & 0.439 & 94.63\% & 92.51\% \\
Mix Amplitude & 0.249 & 0.413 & 90.36\% & 94.87\% & 84.64\% & 84.25\% & 0.780 & 0.439 & 94.65\% & 92.77\% \\
\midrule
\rowcolor{myblue} Gaussian Model & \textbf{0.255} & \textbf{0.418} & \textbf{90.88\%} & \textbf{95.41\%} & \textbf{86.42\%} & \textbf{85.31\%} & \textbf{0.768} & \textbf{0.433} & \textbf{95.07\%} & \textbf{92.96\%} \\
\bottomrule
\end{tabular}}
}
\label{tab: blur modeling}
\vspace{0.1cm}
\end{table*}

\subsection{Analytical Experiments}

\textbf{Analysis of Cross-Subject Ridge Reuse.}
We analyze the two key designs in CSRR in Tab.~\ref{tab: CSRR Ablation}. 
``SourceFuse'' denotes feature fusion with source-subject ridge priors, and ``CosLoss'' denotes the cosine distillation loss that regularizes the target ridge output toward the source-prior representation.
As shown in Tab.~\ref{tab: CSRR Ablation}, using either SourceFuse or CosLoss alone improves most reconstruction metrics over the baseline. 
With only CosLoss, PixCorr increases from $0.248$ to $0.250$, Inception from $85.15\%$ to $86.24\%$, and CLIP from $83.94\%$ to $84.79\%$. 
With only SourceFuse, PixCorr further increases to $0.251$, Alex(2) to $90.88\%$, and Inception to $86.33\%$. 
When both components are used together, CSRR achieves the best overall performance, including PixCorr $0.255$, CLIP $85.31\%$, image retrieval $95.07\%$, and brain retrieval $92.96\%$. 
These results suggest that SourceFuse and CosLoss provide complementary regularization for stabilizing the target-subject projection under limited-data adaptation.

\textbf{Effect of Source-Prior Construction.}
We compare three strategies for constructing the source prior in Tab.~\ref{tab: source prior construct}. 
``Random One'' randomly selects one source-subject ridge as the prior, ``Nearest One'' selects the source subject whose ridge representation is most similar to the target subject, and ``All Average'' averages ridge outputs from all source subjects as our default strategy.
As shown in the table, ``All Average'' achieves the best overall performance, improving Inception from $85.70\%$ to $86.42\%$, CLIP from $84.25\%$ to $85.31\%$, image retrieval from $94.53\%$ to $95.07\%$, and brain retrieval from $92.25\%$ to $92.96\%$ compared with ``Random One''. 
Although ``Nearest One'' obtains comparable PixCorr ($0.255$), it is worse on CLIP ($83.67\%$) and brain retrieval ($92.10\%$). 
These results suggest that aggregating multiple source priors provides more reliable cross-subject guidance than relying on a single source.

\textbf{Effect of Image-Side Blur.}
We evaluate different image-side supervision strategies in Tab.~\ref{tab: blur strategies}. 
``Clean image'' uses the original image without blur, ``Whole blur'' applies uniform blur to the entire image, ``Fixed-radius blur'' preserves a fixed central region while blurring the surrounding area, and ``Difficulty-aware blur'' adaptively adjusts the clear region and blur strength according to sample-level alignment difficulty.
Compared with clean supervision, ``Whole blur'' improves high-level alignment, increasing Inception from $85.17\%$ to $85.68\%$ and CLIP from $83.21\%$ to $84.26\%$, but slightly reduces retrieval performance (image / brain retrieval: $93.84\% / 92.18\%$ vs. $94.13\% / 92.51\%$). 
``Fixed-radius blur'' provides a stronger balance, achieving $0.251$ PixCorr, $84.70\%$ CLIP, and $92.69\%$ brain retrieval. 
The proposed difficulty-aware blur achieves the best overall results, improving PixCorr to $0.255$, Inception to $86.42\%$, CLIP to $85.31\%$, and image / brain retrieval to $95.07\% / 92.96\%$. 
These results suggest that adaptive blur provides more effective image-side supervision than either full-detail supervision or non-adaptive blur.

\begin{table*}[!t]
\centering
\caption{
Ablation on the positions of feature-level brain augmentation (FBA) on Subject~1. 
``Vanilla'' denotes StableMind without FBA, ``Block $k$'' applies FBA to a single backbone block, and ``Blocks $i$-$j$'' applies FBA to multiple blocks.
}
\resizebox{\textwidth}{!}{
      \setlength{\tabcolsep}{2.3mm}{
\begin{tabular}{l|cccc|cccc|cc}
\toprule
\multirow{2}{*}{\shortstack{\textbf{Feature}\\\textbf{Augmentation}}} 
% \multirow{2}{*}{\shortstack{\textbf{FBA}\\\textbf{Variant}}}
& \multicolumn{4}{c|}{\textbf{Low-Level}} & 
\multicolumn{4}{c|}{\textbf{High-Level}} & 
\multicolumn{2}{c}{\textbf{Retrieval}} \\ \cline{2-11} \noalign{\vskip 1.5pt} 
 & PixCorr$\uparrow$ & SSIM$\uparrow$ & Alex(2)$\uparrow$ & Alex(5)$\uparrow$ & Incep$\uparrow$ & CLIP$\uparrow$ & Eff.$\downarrow$ & SwAV$\downarrow$  & Image$\uparrow$ & Brain$\uparrow$ \\ 
\midrule
Vanilla & 0.245 & 0.414 & 89.65\% & 94.50\% & 85.36\% & 84.28\% & 0.783 & 0.440 & 94.71\% & 92.31\% \\
\midrule
Block 1 & 0.245 & 0.416 & 89.99\% & 94.63\% & 84.85\% & 83.81\% & 0.781 & 0.445 & 94.33\% & 92.34\% \\
Block 2 & 0.249 & 0.417 & 90.40\% & 94.92\% & 84.80\% & 84.66\% & 0.778 & 0.439 & 94.77\% & 92.74\% \\
Block 3 & 0.249 & 0.416 & 90.51\% & 95.27\% & 85.92\% & 84.97\% & 0.779 & 0.439 & 94.89\% & 92.38\% \\
Block 4 & 0.241 & 0.399 & 89.23\% & 94.03\% & 83.86\% & 83.43\% & 0.790 & 0.451 & 93.98\% & 91.60\% \\
\midrule
Blocks 1-2 & 0.247 & 0.415 & 90.03\% & 94.87\% & 86.13\% & 84.59\% & 0.775 & 0.439 & 94.59\% & \textbf{93.44\%} \\
Blocks 2-3 & \textbf{0.255} & \textbf{0.418} & \textbf{90.88\%} & \textbf{95.41\%} & \textbf{86.42\%} & \textbf{85.31\%} & \textbf{0.768} & \textbf{0.433} & \textbf{95.07\%} & 92.96\% \\
Blocks 1-3 & 0.249 & 0.415 & 90.29\% & 94.89\% & 85.95\% & 85.10\% & 0.771 & 0.437 & 94.94\% & 93.09\% \\
\bottomrule
\end{tabular}}
}
\label{tab: FBA positions}
% \vspace{0.1cm}
\end{table*}

\begin{figure}[!t]
  \centering
    \includegraphics[width=\linewidth]{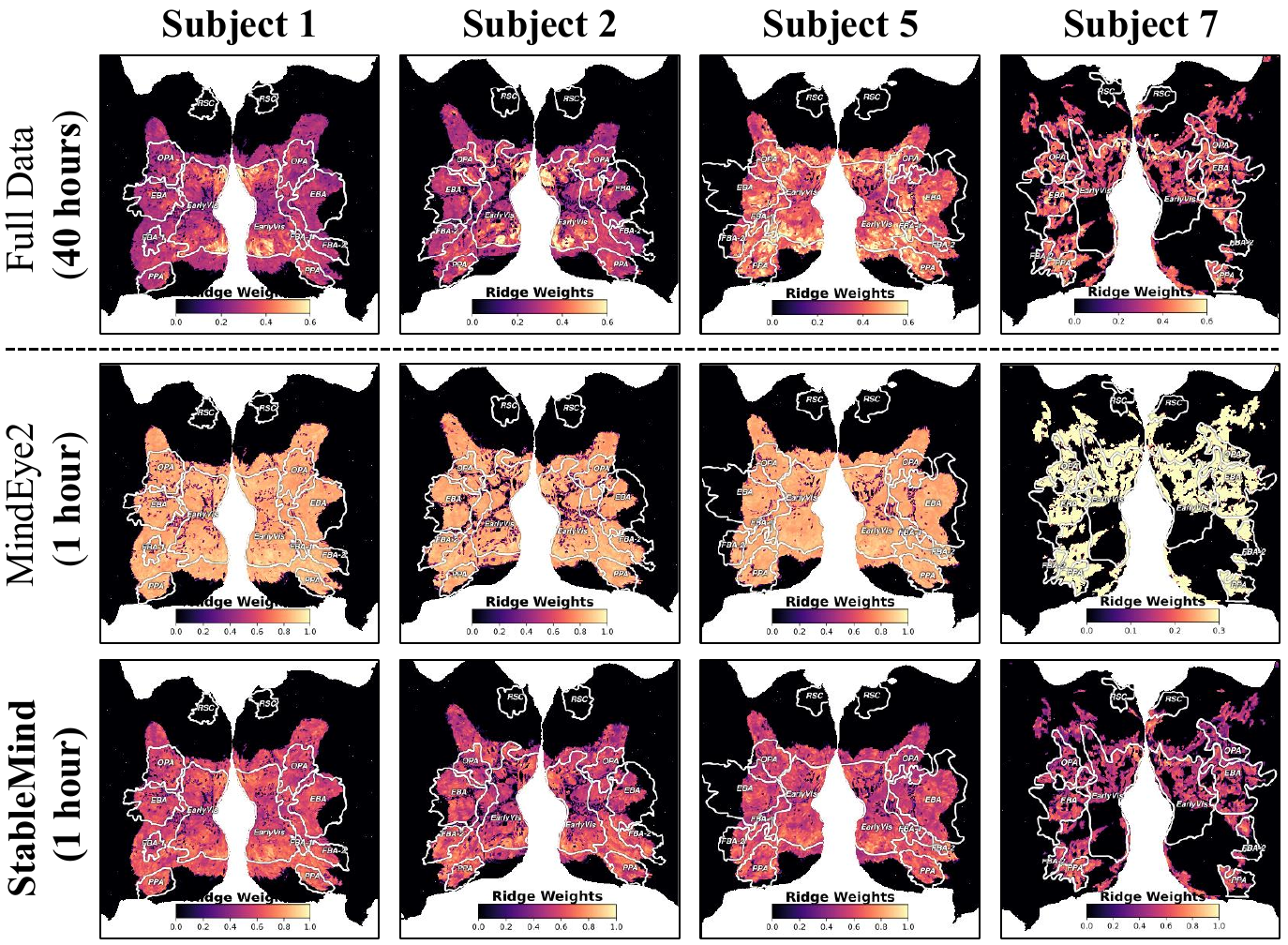}
  \caption{The Voxel-wise ridge weights on the NSD cortical flat map. For each subject, we project the first-layer ridge weights onto the cortical surface using pycortex \cite{gao2015pycortex}. The top row is the model trained on the full 40-hour dataset (as an upper bound), the middle and bottom rows are the MindEye2 and our StableMind trained with only 1 hour of data.
 }
  \label{fig:brain_roi}
\end{figure}

\textbf{Why Use Gaussian Distribution for Amplitude Modeling?}
\label{sec. gaussian modeling}
We compare multiple FBA variants with different augmentations in Tab.~\ref{tab: blur modeling}.
``Vanilla'' denotes StableMind without feature-level brain augmentation. 
``Random Noise'' adds element-wise random noise to the amplitude spectrum. 
``Uniform Model'' samples the mean and standard deviation of amplitude statistics from a uniform distribution. 
``Swap Amplitude'' replaces the amplitude statistics of one sample with those from another sample, while ``Mix Amplitude'' interpolates amplitude statistics between samples. 
``Gaussian Model'' samples amplitude statistics from a Gaussian distribution estimated from mini-batch statistics.
As shown in Tab.~\ref{tab: blur modeling}, unstructured perturbations are less effective. 
``Random Noise'' slightly improves low-level metrics but reduces high-level alignment and retrieval, with CLIP decreasing to $83.27\%$ and brain retrieval to $91.39\%$. 
Similarly, ``Uniform Model'' underperforms Vanilla in CLIP ($83.44\%$ vs. $84.28\%$) and brain retrieval ($91.08\%$ vs. $92.31\%$), indicating that uniformly perturbing amplitude statistics may introduce unrealistic feature shifts. 
``Swap Amplitude'' and ``Mix Amplitude'' provide more structured sample-level perturbations and improve PixCorr ($0.250$ and $0.249$) as well as brain retrieval ($92.51\%$ and $92.77\%$), but their CLIP scores remain lower than Vanilla. 
In contrast, ``Gaussian Model'' achieves the best overall performance, improving PixCorr to $0.255$, CLIP to $85.31\%$, and brain retrieval to $92.96\%$.
This suggests that modeling amplitude perturbations as smooth statistical shifts around the empirical distribution provides a more effective regularization than unstructured or distribution-agnostic perturbations.
Thus, Gaussian amplitude modeling provides an effective feature-level augmentation for improving robustness to subject-specific fMRI variations while preserving high-level semantic alignment.

\textbf{Different Positions of Feature-level Brain Augmentation.}
As shown in Tab.~\ref{tab: FBA positions}, we evaluate the effectiveness of feature-level brain augmentation (FBA) at different blocks of the backbone (containing four blocks). 
The results show that FBA is most effective when applied to intermediate blocks (Blocks 2 or 3), while applying it to Block 1 yields moderate improvements. 
In contrast, applying FBA to Block 4 leads to a marginal performance drop. 
One possible explanation is that features at Block 4 are directly used by multiple training objectives, making them more sensitive to perturbations.
Excessive perturbations at this stage could disrupt both reconstruction and retrieval performance.
We further investigate multi-layer configurations of FBA. 
Applying FBA jointly to Blocks 2 and 3 achieves the best overall performance, while extending it to Blocks 1-3 results in slightly inferior results. 
This observation suggests that overly strong perturbations may hinder the model's ability to learn consistent semantic representations. 
Therefore, we adopt Blocks 2 and 3 as the default FBA positions for all cross-subject adaptation experiments.

\begin{figure*}[!t]
  \centering
    \includegraphics[width=\linewidth]{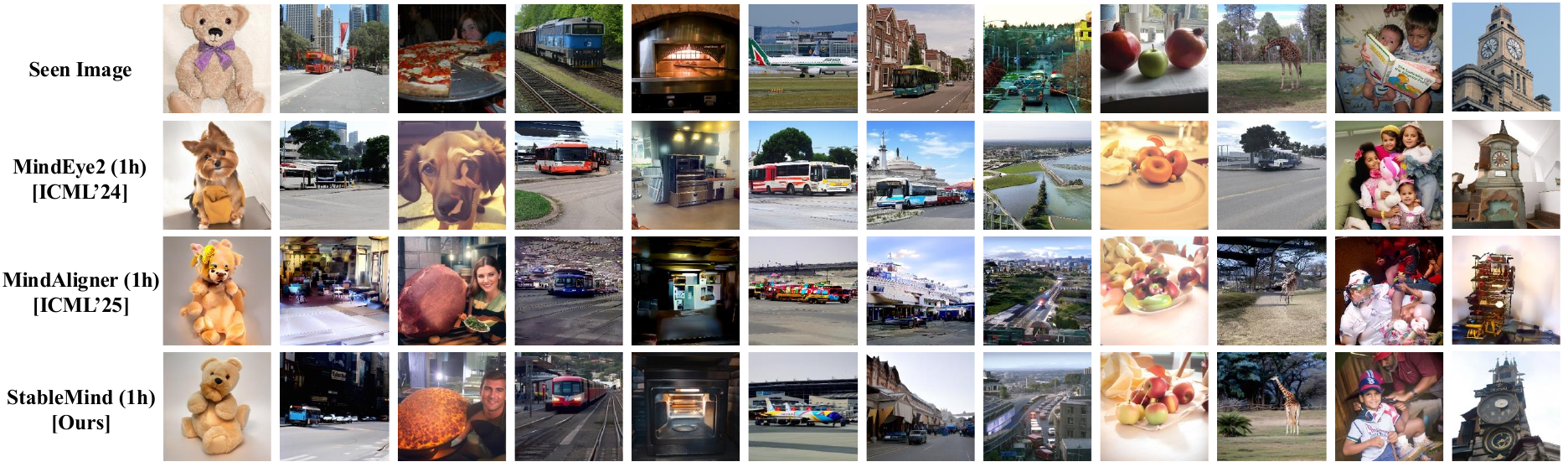}
  \caption{
 Qualitative fMRI-to-image reconstructions of our StableMind and other representive methods, including MindEye2 \cite{scotti2024mindeye2} and MindAligner \cite{dai2025mindaligner}. StableMind produces reconstructions that more closely match the visual stimuli in both appearance and semantics than previous methods.
 }
  \label{fig: recons}
\end{figure*}

\begin{figure}[!t]
  \centering
    \includegraphics[width=\linewidth]{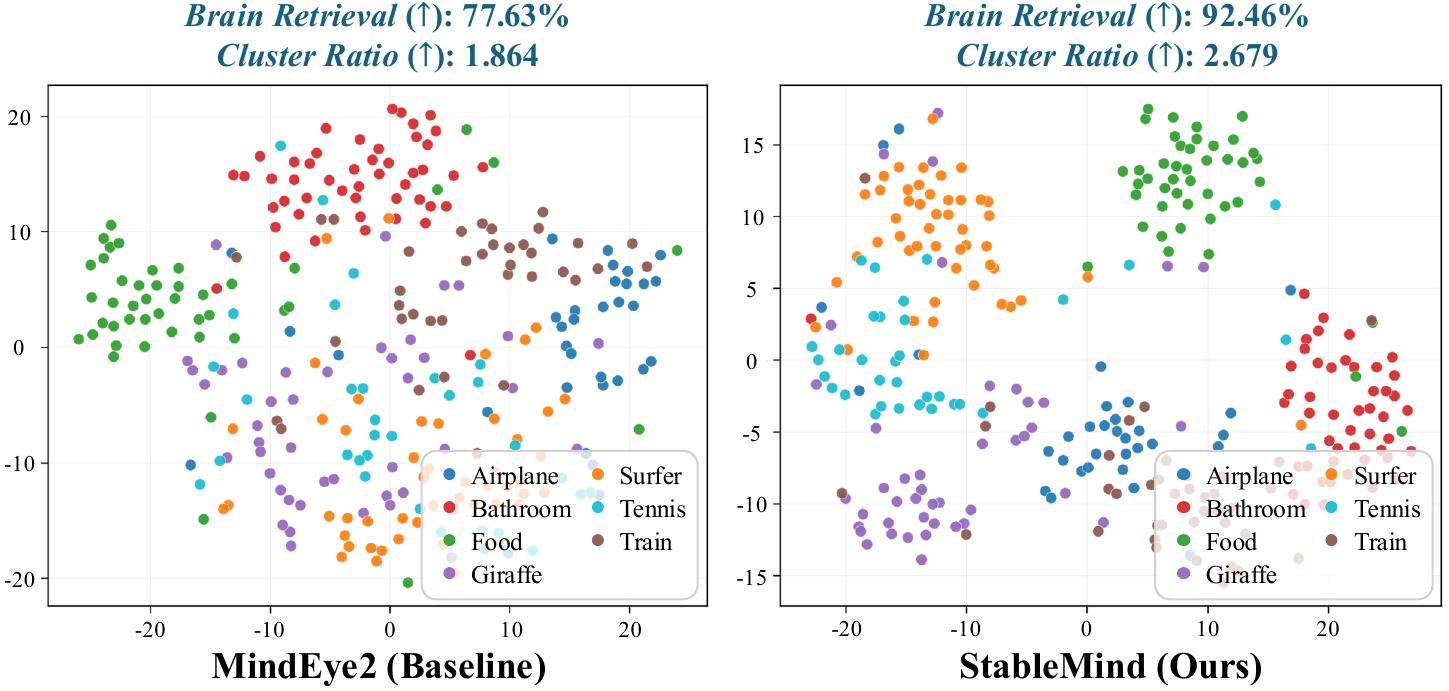}
  \caption{
    The t-SNE visualization of MindEye2 and StableMind finetuned on subject $1$ under the 1-hour adaptation setting. StableMind yields tighter intra-class clusters and clearer inter-class separation than MindEye2. The numbers above each panel denote brain-retrieval accuracy and cluster ratio. }
  \label{fig: tsne comparison}
\end{figure}

\textbf{Neuroscience Interpretability of StableMind.}
Following MindTuner~\cite{gong2025mindtuner}, we further assess where subject-specific structure is captured in the visual cortex.
For each subject, we compare three decoders, including (1) the MindEye2 model finetuned on the full 40-hour NSD training set, which serves as an approximate upper bound; (2) the MindEye2 model trained with only 1 hour of fMRI per subject; and (3) our StableMind model under the same 1-hour budget.
We then use pycortex \cite{gao2015pycortex} to project the first-layer ridge weights onto the NSD 2D flat map and treat the magnitude of each weight as the importance of the corresponding voxel.
As shown in Fig.~\ref{fig:brain_roi}, the full-data ridge model produces structured weight patterns with clear separations between different visual areas.
When trained with only $2.5\%$ of the data, MindEye2 yields more fragmented and low-contrast maps, where different visual regions are poorly differentiated and voxel-wise dependencies are harder to interpret.
In contrast, our StableMind recovers sharper and more coherent patterns that more closely resemble the full-data results, concentrating importance within known visual ROIs.
These results suggest that StableMind learns more spatially coherent voxel-weight patterns under limited-data adaptation, which are more consistent with the full-data ridge maps.

\textbf{Feature Visualization.}
We further visualize the learned representations using t-SNE in Fig.~\ref{fig: tsne comparison}. 
Compared with MindEye2, StableMind produces more compact intra-class clusters and clearer separation between different classes. 
This is reflected by both higher brain-retrieval accuracy and a larger cluster ratio, indicating improved discriminability of the learned representations.
These visualization results are consistent with the quantitative gains brought by the proposed components.
CSRR regularizes the subject-specific projection by leveraging source priors, while FBA introduces structured perturbations that enhance robustness to cross-subject variability. 
In addition, DIB reduces the influence of unstable fine-grained visual details during alignment, leading to more semantically consistent representations. 
As a result, StableMind better preserves class-level structure under limited-data adaptation, which is consistent with its gains in retrieval performance.

\begin{table}[!t]
\centering
\caption{
Cross-subject spectral discrepancy in the CLIP-aligned latent space. 
We report the average pairwise frequency, amplitude, and phase gaps across target subjects. 
Lower values indicate smaller cross-subject discrepancy. 
}
\resizebox{\linewidth}{!}{
\begin{tabular}{l|ccc}
\toprule
\textbf{Method} & Frequency Gap$\downarrow$ & Amplitude Gap$\downarrow$ & Phase Gap$\downarrow$ \\
\midrule
MindEye2 \cite{scotti2024mindeye2} & 40.395 & 20.068 & 0.162  \\
MindAligner \cite{dai2025mindaligner} & 29.826 & 12.337 & 0.162 \\
MindTuner \cite{gong2025mindtuner} & 16.550 & 10.145 & 0.159 \\
\rowcolor{myblue} \textbf{StableMind} {\scriptsize (Ours)} & \textbf{8.449} & \textbf{6.078} & \textbf{0.157} \\
\bottomrule
\end{tabular}
}
\label{tab:frequency_gap}
\end{table}

\textbf{Effectiveness of StableMind in Reducing Cross-Subject Gaps.}
To investigate the effect of StableMind on cross-subject alignment, we quantify spectral discrepancies between subjects in the CLIP-aligned latent space.
For each subject $s$, let the projected representation be $\mathbf{z}^{(s)} \in \mathbb{R}^{B \times 1 \times N}$, where $B$ is the number of samples and $N$ is the feature length.
We compute the 1-D discrete Fourier transform:
\begin{equation}
 \mathcal{F}\!\left(\mathbf{z}^{(s)}\right)[k]
 = \sum_{n=0}^{N-1} \bar{\mathbf{z}}^{(s)}[n]\,
 e^{-j 2\pi nk / N}, \quad k = 0,\dots,N-1,
\end{equation}
and obtain the corresponding amplitude and phase spectra as $\mathcal{A}\!\left(\mathbf{z}^{(s)}\right)$ and $\mathcal{P}\!\left(\mathbf{z}^{(s)}\right)$.
We then average the frequency and amplitude spectra over the batch dimension:
\begin{equation}
 \bar{\mathcal{F}}^{(s)}
 = \frac{1}{B} \sum_{b=1}^{B} \mathcal{F}\!\left(\mathbf{z}_b^{(s)}\right),
\end{equation}
\begin{equation}
 \bar{\mathcal{A}}^{(s)}
 = \frac{1}{B} \sum_{b=1}^{B} \mathcal{A}\!\left(\mathbf{z}_b^{(s)}\right).
\end{equation}
For phase, we use a circular average to respect its periodicity:
\begin{equation}
\bar{\mathcal{P}}^{(s)}
= \operatorname{Angle}\!\left(
\frac{1}{B} \sum_{b=1}^{B} e^{\,j \mathcal{P}(\mathbf{z}_b^{(s)})}
\right).
\end{equation}
For a pair of subjects $(p,q)$, the frequency, amplitude, and phase gaps are computed as:
\begin{equation}
D_{\mathrm{freq}}(p,q)
 = \bigl\|
 \bar{\mathcal{F}}^{(p)}
 -
 \bar{\mathcal{F}}^{(q)}
 \bigr\|_{2},
\end{equation}
\begin{equation}
D_{\mathrm{amp}}(p,q)
 = \bigl\|
 \bar{\mathcal{A}}^{(p)}
 -
 \bar{\mathcal{A}}^{(q)}
 \bigr\|_{2},
\end{equation}
\begin{equation}
  D_{\text{pha}}(p,q)=
\left\|
\operatorname{Angle}
\left(
e^{j(\bar{\mathcal{P}}^{(p)}-\bar{\mathcal{P}}^{(q)})}
\right)
\right\|_2 .
\end{equation}
The values in Tab.~\ref{tab:frequency_gap} are obtained by averaging these pairwise gaps over all subject pairs.
As shown in Tab.~\ref{tab:frequency_gap}, StableMind achieves the lowest spectral discrepancies among all compared methods. 
Compared with MindEye2, StableMind reduces the frequency gap from $40.395$ to $8.449$ and the amplitude gap from $20.068$ to $6.078$. 
It obtains lower gaps than MindAligner and MindTuner, with an amplitude gap of $6.078$ compared with $12.337$ and $10.145$, respectively. 
The phase gap is relatively small for all methods, while StableMind still achieves the lowest value ($0.157$). 
These results indicate that StableMind extracts consistent latent representations across subjects, especially in amplitude-related spectral statistics, consistent with the motivation of feature-level brain augmentation.

\textbf{Reconstruction Results.} 
We visualize fMRI-to-image reconstructions to qualitatively assess the effectiveness of our framework.
As illustrated in Fig.~\ref{fig: recons}, we treat Subject~1 as a new subject and fine-tune the models on 1-hour data of Subject~1.
The first row shows the visual stimuli, and the second row shows reconstructions from a decoder trained on the full 40-hour data.
The third and fourth rows correspond to MindEye2 \cite{scotti2024mindeye2} and MindAligner \cite{dai2025mindaligner} trained with 1 hour of fMRI data for Subject~1, while the last row reports the reconstructions from our StableMind under the same 1-hour budget.
Compared with other methods, StableMind tends to recover object shapes, colors, and scene layouts that are closer to the visual stimuli.
The results indicate that StableMind effectively exploits limited fMRI data to reconstruct meaningful visual content.

\section{Conclusion}

In this work, we presented StableMind, a regularized source-free adaptation framework for cross-subject fMRI decoding under limited target-subject data.
Rather than focusing solely on latent alignment, our method explicitly addresses two key challenges in practical adaptation, including the strong subject-specific variability in brain signals and the mismatch between brain responses and fine-grained visual supervision.
To tackle these issues, StableMind introduces a set of complementary designs that operate at different stages of the adaptation process. 
We first regularize the voxel-to-latent projection by incorporating cross-subject ridge priors, which stabilize target-subject mapping under limited data.
We then improve representation robustness via Fourier-based feature-level brain augmentation, which reduces sensitivity to subject-dependent variations. 
Finally, we refine the image-side supervision through a difficulty-aware blur strategy, which mitigates the influence of unreliable visual details during alignment.
Experiments on the NSD benchmark under the 1-hour setting show that StableMind improves retrieval performance and achieves competitive reconstruction quality against strong baselines, while requiring fewer trainable adaptation parameters.
Further analyses indicate that the proposed design effectively reduces cross-subject spectral discrepancies and leads to structured latent representations. 
Our results suggest that improving cross-subject brain decoding requires not only better alignment, but also effective adaptation regularization at both the subject-transfer and stimulus-supervision levels.

\bibliographystyle{IEEEtran}
\bibliography{IEEEfull}

% \vspace{-1.0cm}

\end{document}